\documentclass[lettersize,journal]{IEEEtran}
\usepackage[table]{xcolor}
\usepackage{amsmath,amsfonts}
\usepackage{algorithmic}
\usepackage{algorithm}
\usepackage{array}
\usepackage[caption=false,font=normalsize,labelfont=sf,textfont=sf]{subfig}
\usepackage{textcomp}
\usepackage{stfloats}
\usepackage{url}
\usepackage{verbatim}
\usepackage{graphicx}
\usepackage{cite}
\usepackage{orcidlink}
\usepackage{float}  
\usepackage{multirow}
\usepackage{adjustbox}
\hypersetup{hidelinks}
\usepackage{overpic}   
\hypersetup{
  colorlinks   = true, 
  urlcolor     = blue, 
  linkcolor    = blue, 
  citecolor   = blue 
}
\renewcommand{\tablename}{Table}

\newcommand{\RCC}[1]{\textcolor{black}{ #1}}

\newcommand{\modulename}{\textcolor{black}{Edge-to-Edge-Node-to-Node Graph Convolution (E2E-N2N-GCN)}}

\begin{document}

\title{Unified Spatial-Temporal Edge-Enhanced Graph Networks for Pedestrian Trajectory Prediction}

\author{Ruochen Li$^{\orcidlink{0000-0001-8966-9613}}$, 
Tanqiu Qiao$^{\orcidlink{0000-0002-6548-0514}}$,
Stamos Katsigiannis$^{\orcidlink{0000-0001-9190-0941}}$,~\IEEEmembership{Member,~IEEE}, 
Zhanxing Zhu$^{\orcidlink{0000-0002-2141-6553}}$, 

Hubert P. H. Shum$^{\orcidlink{0000-0001-5651-6039}\dag}$,~\IEEEmembership{Senior Member,~IEEE}

\thanks{R. Li, T. Qiao, S. Katasigiannis and H. P. H. Shum are with Durham University, UK.
        (e-mail: \{ruochen.li, 
        tanqiu.qiao,
        stamos.katsigiannis, hubert.shum\}@durham.ac.uk).}
\thanks{Z. Zhu is with the University of Southampton, UK. (e-mail: z.zhu@soton.ac.uk)}
\thanks{\dag Corresponding author: H. P. H. Shum}
}

\maketitle

Pedestrian trajectory prediction aims to forecast future movements based on historical paths. Spatial-temporal (ST) methods often separately model spatial interactions among pedestrians and temporal dependencies of individuals. They overlook the direct impacts of interactions among different pedestrians across various time steps (i.e., high-order cross-time interactions). This limits their ability to capture ST inter-dependencies and hinders prediction performance. To address these limitations, we propose UniEdge with three major designs. Firstly, we introduce a unified ST graph data structure that simplifies high-order cross-time interactions into first-order relationships, enabling the learning of ST inter-dependencies in a single step. This avoids the information loss caused by multi-step aggregation. 
Secondly, traditional GNNs focus on aggregating pedestrian node features, neglecting the propagation of implicit interaction patterns encoded in edge features. We propose the Edge-to-Edge-Node-to-Node Graph Convolution (E2E-N2N-GCN), a novel dual-graph network that jointly models explicit N2N social interactions among pedestrians and implicit E2E influence propagation across these interaction patterns.
Finally, to overcome the limited receptive fields and challenges in capturing long-range dependencies of auto-regressive architectures, we introduce a transformer encoder-based predictor that enables global modeling of temporal correlation. UniEdge outperforms state-of-the-arts on multiple datasets, including ETH, UCY, and SDD.


\section{Introduction}

\begin{figure}[t]\centering
  \includegraphics[width=\linewidth]{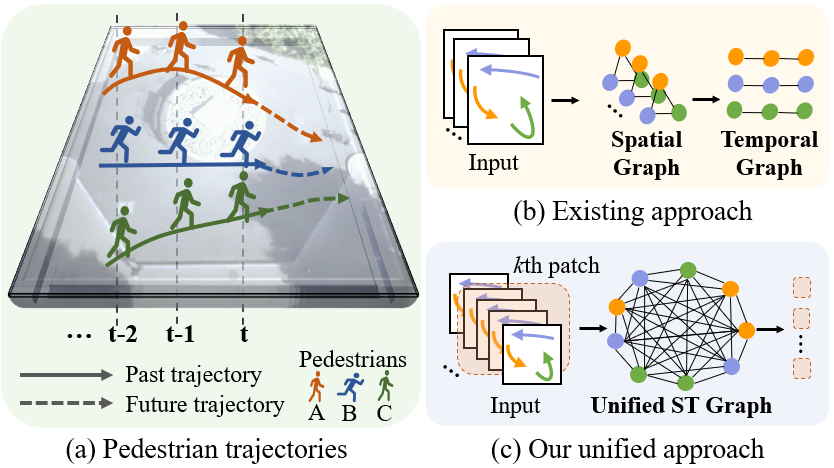}

  \caption{Motivation Illustration. \textbf{(a)} \textbf{Real-world pedestrian trajectories} over multiple time frames.
  \textbf{(b)} \textbf{Existing ST approaches} separately model the spatial interactions among pedestrians and temporal dependencies of individuals.
  \textbf{(c)} \textbf{Our unified ST graph} integrates ST inter-dependencies and simplifies high-order cross-time interactions into first-order relationships.}
  \label{fig:teaser}
\end{figure}

\IEEEPARstart{T}{he} \RCC{aim of pedestrian trajectory prediction is to forecast future paths based on observed movements (\figurename~\ref{fig:teaser}(a)). High-precision prediction systems are crucial for applications like self-driving vehicles \cite{bai2015pomdpintro, chen2023ppnet} and video surveillance \cite{sun2021reciprocaltrajectory}. Specifically, in intelligent surveillance systems, especially at accident-prone intersections, early detection of pedestrian crossing intentions within a few seconds enables timely warnings to approaching vehicles through Vehicle-to-Everything (V2X) communication between vehicles, infrastructure and pedestrians, providing sufficient time for vehicles to react and reduce accident risks \cite{zhou2022realapp}.}

Predicting pedestrian trajectory is inherently challenging, primarily due to the complexity of interactions in which pedestrians continuously adjust their movements based on the evolving dynamics of others over multiple time steps. Spatial-temporal (ST) graph architectures (\figurename~\ref{fig:teaser}(b)) are widely used to analyze human motions \cite{liu2020trajectorycnn, wang2023stmotion} and pedestrian trajectories \cite{shi2021sgcn,ruochen2022multiclassSGCN,kosaraju2019socialbigat,huang2019stgat,Mohamed2020socialstgcnn, bae2022gpgraph, bae2023eigentrajectory, bae2023TERN}, capturing spatial interactions within each frame and temporal dependencies over time.

\RCC{This challenge is particularly severe when modeling \textbf{high-order cross-time interactions}, i.e., complex interactions among pedestrians across multiple time steps. Traditional ST graph architectures require multiple steps to capture these interactions, where each node first aggregates spatial information at individual time steps and then addresses temporal dependencies through temporal networks. STGAT \cite{huang2019stgat} combines graph attention \cite{petar2017GAT} with Long Short-Term Memory (LSTM) \cite{HochSchm1997lstm} for sequential temporal modeling, while Social-STGCNN \cite{Mohamed2020socialstgcnn} and SGCN \cite{shi2021sgcn} advance to integrating Graph Convolutional Network (GCN) \cite{kipf2016GCN} with Temporal Convolutional Network (TCN) \cite{bai2018TCN} for parallel processing. This paradigm has two key disadvantages: (1) when processing high-order interactions among pedestrians, this multi-step aggregation paradigm leads to potential under-reaching \cite{lu2024nodemixup} due to increased effective resistance \cite{black2023understandinger}, where important interaction patterns are diluted and compressed with the increase of aggregation steps; and (2) the separation of spatial and temporal processing can disrupt the natural unified ST inter-dependencies observed in real-world scenarios \cite{Wang2023FullyConnectedSG, yi2023fouriergnn}, particularly in situations requiring immediate response to dynamic changes.}

\begin{figure}[t]\centering
  \includegraphics[width=0.8\linewidth]{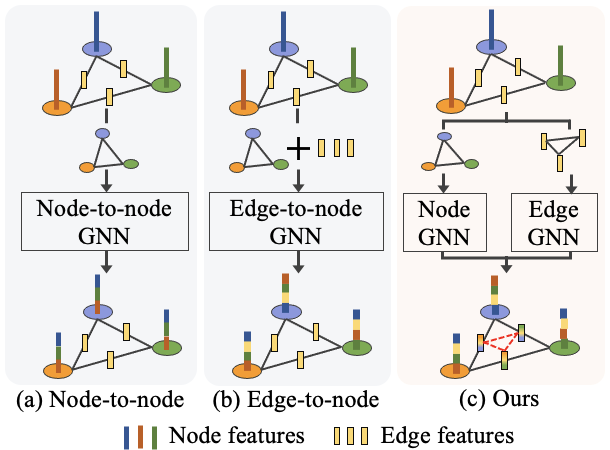}
  \caption{\RCC{Illustration of graph learning procedures. (a) Node-to-Node (N2N), (b) Edge-to-Node (E2N), and (c) Our novel dual-graph introduces the combination of N2N and Edge-to-Edge (E2E) paradigm.}}
  \label{fig:graph_type}
\end{figure}

\RCC{Another challenge lies in modeling the implicit influence propagation through edges in pedestrian social interactions. While Graph Neural Networks (GNNs) are widely adopted for modeling pedestrian interactions \cite{huang2019stgat, Mohamed2020socialstgcnn, bae2022gpgraph}, existing approaches primarily focus on \textbf{Node-to-Node} (N2N) interactions (\figurename~\ref{fig:graph_type}(a)) through GNNs, e.g., using inverse distance \cite{Mohamed2020socialstgcnn} or attention-based \cite{shi2021sgcn,huang2019stgat} weighting. Recent works like GroupNet \cite{Xu2022GroupNetMH} and HEAT \cite{mo2021edge_mask} advance to \textbf{Edge-to-Node} (E2N) interactions (\figurename~\ref{fig:graph_type}(b)) by incorporating edge features into node representations, enhancing the relation reasoning ability of the system. However, both N2N and E2N focus on the training of node features, while neglecting the crucial \textbf{Edge-to-Edge} (E2E) patterns \cite{xia2023deciphering, huang2023brainfunction}. This fundamental limitation restricts GNNs' ability to capture the full spectrum of interaction dynamics in pedestrian behaviors, particularly in complex ST scenarios where one pedestrian's behavior can implicitly influence others through cascade effects \cite{xia2023deciphering}.}


\RCC{In this paper, we introduce the Unified Spatial-Temporal Edge-enhanced Graph Network (UniEdge) for pedestrian trajectory prediction. To address the first challenge, our unified ST graph segments input trajectories into patch-based structures (\figurename~\ref{fig:teaser}~(c)), simplifying high-order cross-time interactions into first-order relationships. This approach reduces effective resistance \cite{black2023understandinger} and mitigates the under-reaching problem \cite{lu2024nodemixup}, preventing information dilution during propagation. By processing ST information jointly in a single step, each unified patch maintains natural ST inter-dependencies, enabling immediate responses to dynamic changes while preserving multi-step interaction patterns.}

\RCC{To tackle the second challenge, we introduce Edge-to-Edge-Node-to-Node Graph Convolution (E2E-N2N-GCN), a dual-graph network that jointly processes both node and edge patterns, as depicted in \figurename~\ref{fig:graph_type}~(c). Dual-graph design provides a deeper understanding of graph topology in various domains \cite{wu2024deepdual, huang2023brainfunction}. Our dual-graph architecture consists of two complementary graphs: a node-level graph that models explicit N2N social interactions among pedestrians, and an edge-level graph that captures the implicit E2E influence propagation across these interaction patterns. Specifically, we employ a first-order boundary operator \cite{post2007firstboundary} to construct edge graphs that reveal how interaction patterns influence each other through connected edges. This approach enables nuanced analysis of both individual behaviors and collective dynamics, essential for predictive accuracy in crowded environments.}

\RCC{Finally, we introduce a Transformer encoder-based predictor to overcome the limited receptive fields and long-range dependency challenges of auto-regressive architectures. Our predictor leverages attention mechanisms \cite{vaswani2017transformer} to enable global modeling of temporal correlations through learnable placeholders, substantially improving the prediction capability.}

Our approach outperforms state-of-the-art methods on commonly used pedestrian trajectory prediction datasets, including ETH \cite{pellegrini2009ETH}, UCY \cite{Lerner2007UCY} and Stanford Drone Dataset (SDD) \cite{Robicquet2016SDD}. The source code for UniEge is openly released on  \url{https://github.com/Carrotsniper/UniEdge}.

Our contributions can be summarized as follows:
\begin{itemize}
      \item We propose a unified ST graph data structure that simplifies high-order cross-time interactions into first-order relationships. This enables direct learning of ST inter-dependencies in a single step, avoiding information loss caused by multi-step aggregation while preserving critical interaction patterns.

      \item We introduce the \modulename, a novel dual-graph architecture that jointly captures both explicit N2N social interactions among pedestrians and implicit E2E influence propagation across interaction patterns through first-order boundary operators. This enables more comprehensive modeling of complex pedestrian behaviors.

    \item We introduce a transformer-based predictor that overcomes the limited receptive fields and challenges associated with capturing long-range dependencies inherent in auto-regressive architectures. This enables global modeling of temporal correlations, substantially improving prediction performance.
\end{itemize}

\section{Related Work}

\subsection{Spatial-Temporal Modeling for Trajectory Prediction}
Spatial-temporal architecture is widely used in trajectory prediction which considers both spatial interactions and temporal dependencies. Pioneering methods such as Social-LSTM \cite{Alexandre2016lstm} and Social-GAN \cite{gupta2018socialgan} propose pooling window mechanisms to compute pedestrian spatial interactions and Long Short-Term Memory (LSTM) \cite{HochSchm1997lstm} for temporal aggregation. Due to the outstanding performance of graphs in representation learning, they are widely used to represent pedestrian interactions. STGAT \cite{huang2019stgat} and Social-BiGAT \cite{kosaraju2019socialbigat} employ Graph Attention Network (GAT) \cite{petar2017GAT} to measure interactions strength and LSTM to capture temporal dependencies. Social-STGCNN \cite{Mohamed2020socialstgcnn} proposes to use a Graph Convolutional Network (GCN) \cite{kipf2016GCN} combined with the TCN \cite{bai2018TCN} to model pedestrian trajectories. To simplify the complexity of the graph, sparse GCN-based approaches \cite{shi2021sgcn, bae2023eigentrajectory, ruochen2022multiclassSGCN} further propose directed graphs to dynamically update graph topology during message passing, and TCN is used to learn temporal correlations. In recent years, group-wise methods \cite{Xu2022GroupNetMH, bae2022gpgraph} have garnered attention due to their superior capability in analyzing group behaviors. 

However, these methods characterize spatial interactions and temporal dependencies separately, leading to diluted information and delayed responses in complex scenarios. To this end, we introduce unified ST graphs that transform high-order interactions into simplified first-order relationships, efficiently capture ST inter-dependencies.

\subsection{Graph Neural Networks}
Graph Neural Networks (GNNs) have gained considerable traction in computer vision tasks due to their ability to model complex relationships and interactions between entities. Harnessing their representational power, GNNs have been successfully applied across various domains, including human skeleton analysis \cite{qiao20222ggcn, yan2018spatial, liu2023skeletonrecognition}, drug design \cite{li2022graphdrug2}, and recommendation systems \cite{wang2019rs1}. In the trajectory prediction domain, the evolution of GNN architectures reflects increasingly sophisticated approaches to modeling social interactions. Early works \cite{Mohamed2020socialstgcnn, yu2018spatiotraffic} primarily relied on the representation capabilities of GCN to model social interactions. Following the success of the self-attention mechanism \cite{vaswani2017transformer}, subsequent studies \cite{kosaraju2019socialbigat, huang2019stgat, shi2021sgcn, ruochen2022multiclassSGCN, shi2023TUTR} enhanced this N2N paradigm by incorporating attention-based GNNs, enabling more adaptive and context-aware relationship modeling. Recent works have begun exploring E2N interactions to capture richer relational information between edge and node. GroupNet \cite{Xu2022GroupNetMH} pioneered this direction by introducing interaction strength and category features to enhance edge significance beyond simple connections. Following this trend, GC-VRNN \cite{xu2023gcvrnn}, HEAT \cite{mo2021edge_mask}, and MFAN \cite{li2024mfan} further advanced E2N modeling by integrating edge features into node embeddings, enhancing relational reasoning capabilities. 

However, existing trajectory prediction methods primarily focus on updating node representations. In this paper, we introduce \modulename, a dual-graph architecture that jointly captures both explicit N2N social interactions and implicit E2E influence propagation, providing a more comprehensive modeling of social interactions.

\subsection{Trajectory Predictor}
\RCC{Trajectory prediction has seen various architectural developments. Early RNN-based approaches \cite{Alexandre2016lstm, gupta2018socialgan, pei2022socialVAE,pei2024autofocusing,sun2021reciprocaltrajectory, Berenguer2021contextual} process trajectories sequentially through hidden states. Among these methods, Social-LSTM \cite{Alexandre2016lstm} processes trajectories where hidden states are updated recursively to capture temporal patterns. Recent works like Social-VAE \cite{pei2022socialVAE} and ATP-VAE \cite{pei2024autofocusing} combine RNN with variational autoencoders to model the uncertainty in trajectory predictions, achieving promising results. Subsequently, TCN-based predictor \cite{Mohamed2020socialstgcnn, shi2021sgcn, ruochen2022multiclassSGCN, li2024mfan} emerged as an alternative approach. Social-STGCNN \cite{Mohamed2020socialstgcnn} combines graph convolutions with TCN to achieve efficient parallel processing through increased receptive fields. SGCN \cite{shi2021sgcn} further advances this design by introducing sparse attention mechanisms to adaptively aggregate temporal features. Recently, transformer-based methods \cite{shi2023TUTR, peng2024mrgtraj, chen2023ppnet} have gained prominence in trajectory prediction, where self-attention mechanisms compute pairwise interactions between all time steps, enabling global temporal modeling without the constraints of sequential processing or fixed receptive fields.}

\RCC{However, RNNs suffer from long-term dependencies due to their auto-regressive nature, and TCNs are limited by fixed receptive fields due to their convolutional structure, while full transformer models have high computational costs. To balance modeling capability and efficiency, we propose a Transformer encoder-based predictor that learns global dependencies within the sequence without high computational costs.}

\begin{figure*}[h]
  \centering
  \includegraphics[width=0.84\textwidth]{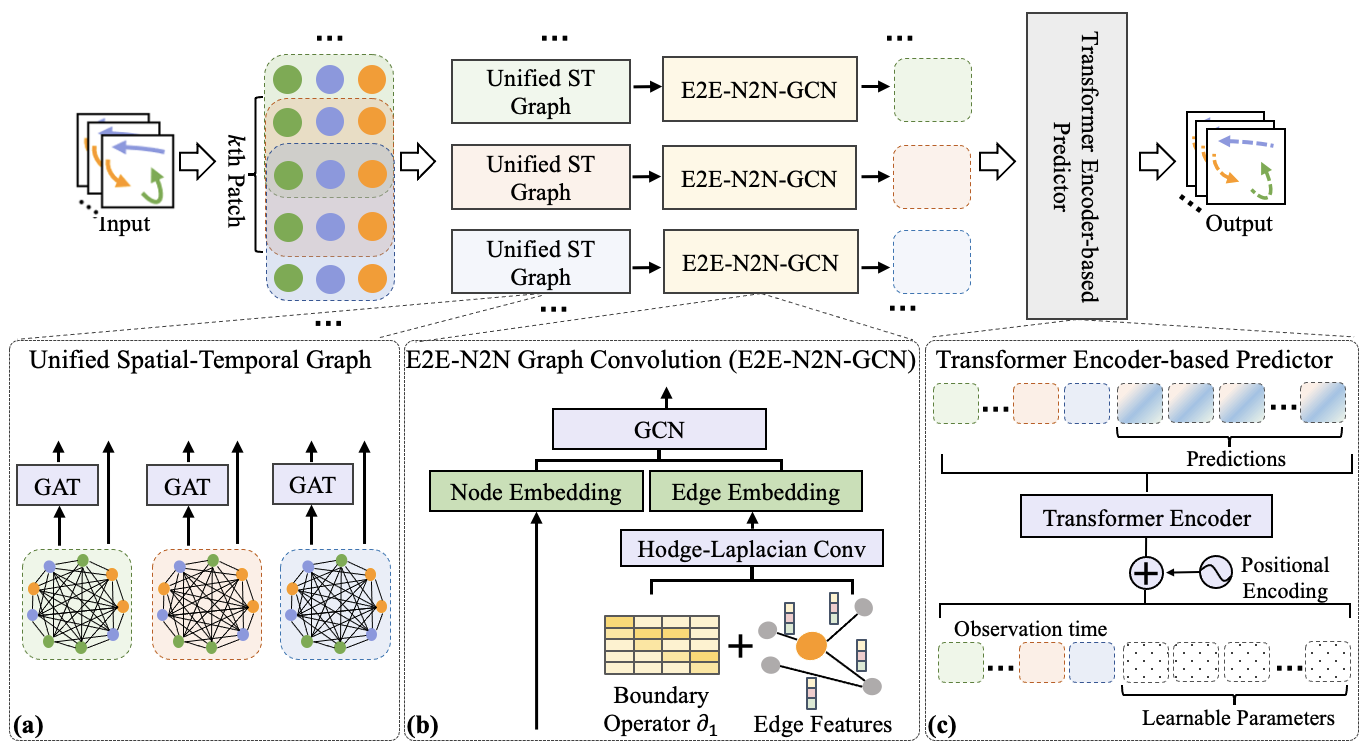}
  \caption{\RCC{Overview of the proposed UniEdge. (a) Construction of patch-based unified ST graphs that simplify cross-time interactions into first-order relationships, (b) Edge-to-Edge-Node-to-Node Graph Convolution (E2E-N2N-GCN) that jointly processes N2N interactions and E2E influence propagation, and (c) Transformer Encoder-based trajectory predictor.}}
  \label{fig:overview}
\end{figure*}

\section{Methodology}
\subsection{Problem Formulation and Feature Initialization}
The goal of pedestrian trajectory prediction is to estimate the possible future trajectories of a pedestrian based on observed trajectories and nearby neighbors. Mathematically, consider a multi-pedestrian scenario containing $N$ pedestrians in  $T_{obs}$ time steps. The observed trajectories of each pedestrian $i \in [1,\dots,N]$ can be represented as $\textit{X}_{i} = \left \{ (x_{t}^{i}, y_{t}^{i}) \mid t\in [-T_{obs}+1,\dots, 0]\right \}$ and its ground-truth future trajectories can be defined as $\textit{Y}_{i} = \left \{ (x_{t}^{i}, y_{t}^{i}) \mid t\in [1,\dots, T_{pred}]\right \}$. For $N$ pedestrians, the observed and ground-truth future trajectories are $\mathbf{X} = [\textit{X}_{1}, \textit{X}_{2}, \dots, \textit{X}_{N}] \in \mathbb{R}^{N \times T_{obs} \times 2}$ and $\mathbf{Y} = [\textit{Y}_{1}, \textit{Y}_{2}, \dots, \textit{Y}_{N}] \in \mathbb{R}^{N \times T_{pred} \times 2}$ respectively, where $2$ denotes the 2D coordinates. Our proposed UniEdge aims to learn a prediction function $\mathcal{F}_{pred}(\cdot)$ that minimizes the differences between the predicted trajectories $\hat{\textbf{Y}} = \mathcal{F}_{pred}(X)$ and the ground-truth future trajectories $\mathbf{Y}$. Instead of directly predicting absolute coordinates, we follow \cite{shi2021sgcn, bae2023eigentrajectory, bae2022gpgraph, Mohamed2020socialstgcnn} that predict relative coordinates of each pedestrian to ensure the robustness and generalization ability of the system across different scenarios.

For trajectory feature initialization, our model takes inputs consisting of pedestrian velocities $\mathbf{v}$, velocity norms $\rho = \|\mathbf{v}\|_{2}$, and pedestrian movement angles $\theta = \text{angle}(\mathbf{v})$, where $\|\cdot\|_{2}$ denotes the vector 2-norm and $\text{angle}(\cdot)$ is the function that computes the angle of the velocity vectors. We follow \cite{wang2024pedestrianICRA} that subtract each historical $\mathbf{v}_{t}, t \in [-T_{obs}, 0]$ by the corresponding endpoint $\mathbf{v}_{T_{pred}}$ as the pre-process step.
These motion dynamic features are embedded and then concatenated to obtain the final geometric feature representation as follows:
$$\mathcal{X} = \text{CONCAT}(f(\mathbf{v}, W_{v}), f(\rho, W_{norm}), f(\theta, W_{angle})),$$
where $\mathcal{X} \in \mathbb{R}^{N \times T_{obs} \times D}$, $N$ and $T_{obs}$ represent the total number of pedestrians and time steps, respectively, and $D$ denotes the embedded feature dimension. Here, $f(\cdot,\cdot)$ represents Multi-Layer Perceptron (MLP) for feature embedding, and $W$ represents the corresponding weights.


\subsection{Unified ST Graph}

Previous trajectory prediction methods often adopt a two-step approach, separately modeling pedestrian spatial interactions and individual temporal dependencies \cite{Alexandre2016lstm,Mohamed2020socialstgcnn,shi2021sgcn}.  This approach is limited in capturing high-order cross-time interactions, which require multi-step aggregation. Such multi-step processing increases the effective resistance - a measurement of graph connectivity that quantifies the efficiency of information flow between nodes \cite{ghosh2008minimizing, black2023understandinger}. High effective resistance impedes graph message-passing, leading to under-reaching problem \cite{lu2024nodemixup}, where message flows from distant nodes are diluted and compressed.

\begin{figure}[t]
  \centering
  \includegraphics[width=0.70\linewidth]{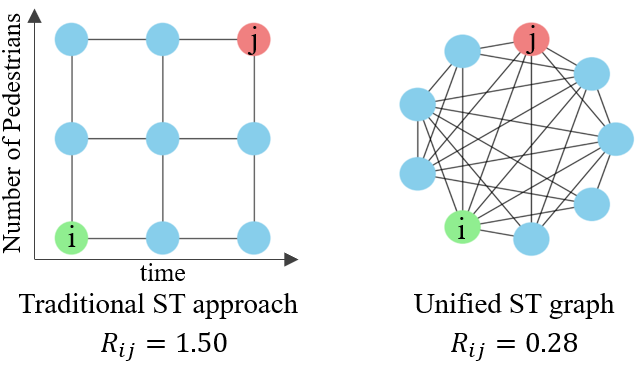}
  \vspace{-3pt}
  \caption{\RCC{Comparison of effective resistance $(R_{ij})$ between traditional ST approach (left, $R_{ij} = 1.50$) and our unified ST graph (right, $R_{ij} = 0.27$). Lower $R_{ij}$ indicates better message propagation efficiency.}}
  \label{fig:effective_resistance}
  \vspace{-0.3cm}
\end{figure}

To address these challenges, we propose a unified ST graph to simplify high-order cross-time interactions among pedestrians into first-order relationships, enabling direct learning of ST inter-dependencies, and preserving high-order interactions without information dilution. This design significantly reduces the effective resistance during message passing, improving information flow efficiency \cite{black2023understandinger, ghosh2008minimizing} and alleviating the risk of under-reaching \cite{lu2024nodemixup}. \figurename~\ref{fig:effective_resistance} illustrates the difference in effective $R$ between the message-passing paradigms of traditional ST approach and our unified approach:
\begin{equation}
R_{ij} = (e_i - e_j)^T L^+ (e_i - e_j)
\end{equation}
where \(L^+\) denotes the Moore-Penrose pseudoinverse of the graph Laplacian matrix representing the graph connectivity \cite{bozzo2013moore}, and \(e_i\), \(e_j\) are standard basis vectors corresponding to nodes \(i\) and \(j\). Lower \(R_{ij}\) values indicate better message propagation efficiency between nodes.

To reduce computational overhead in processing entire sequences and to better capture fine-grained pedestrian dynamics, we adopt a patch-based strategy akin to the local receptive fields used in convolution kernel for image processing. \cite{krizhevsky2012imagenet}.
Specifically, to construct the unified ST graph depicted in \figurename~\ref{fig:overview}~(a), the input features are segmented into $K$ overlapping patches across the temporal dimension $T_{obs}$. These patches are defined by a length $L$ and a stride $\mathcal{S}$, yielding $K = \left\lfloor \frac{T_{obs} - L}{\mathcal{S}} \right\rfloor + 1$. For each patch $k$, ranging from $1$ to $K$, a graph $\mathcal{G}_{node}^{k} = (\mathcal{Z}^{k}, \mathcal{A}_{node}^{k})$ is constructed. Here, $\mathcal{Z}^{k} \in \mathbb{R}^{NL \times D}$ represents the node features, and $\mathcal{A}_{node}^{k} \in \mathbb{R}^{NL \times NL}$ denotes the node adjacency matrix, which encapsulates the node connections. This configuration further benefits subsequent trajectory prediction phases by reducing the number of input tokens from $T_{obs}$ to $K$, which is crucial when using the transformer encoder model. It leads to a quadratic reduction in memory usage and computational complexity for the attention map, by a factor of $ \left(\frac{T_{obs}}{K}\right)^2$. 

We then apply GAT \cite{brody2021gatv2, huang2019stgat, kosaraju2019socialbigat} to initialize interactions strength for the $k$th graph $\mathcal{G}^{k}$ as:
\begin{equation}
\mathcal{H}_{node}^{k} = \text{GAT}(\mathcal{Z}^{k}, \mathcal{A}_{node}^{k}),
\end{equation}
where each node $\mathcal{H}_{node, i}^{k}$ is embedded as:
\begin{equation}
\mathcal{H}_{node, i}^{k} = \sigma\left(\sum_{j \in \mathcal{N}(i) \cup \{i\}} \alpha_{i,j}^{k} \mathbf{\Theta} \mathcal{Z}_{j}^{k} \right),
\end{equation}

\begin{equation}
\alpha_{i,j}^{k} = \frac{\exp\left(\mathbf{a}^{\top}\mathrm{\Gamma}\left(\mathbf{\Theta} [\mathcal{Z}_{i}^{k} \, \Vert \, \mathcal{Z}_{j}^{k}]\right)\right)}{\sum_{j \in \mathcal{N}(i) \cup \{i\}} \exp\left(\mathbf{a}^{\top}\mathrm{\Gamma}\left(\mathbf{\Theta} [\mathcal{Z}_{i}^{k} \, \Vert \, \mathcal{Z}_{j}^{k}]\right)\right)},
\end{equation}
where $\mathbf{\Theta(\cdot)}$ is transformation function, $\Gamma(\cdot)$ and $\sigma(\cdot)$ denote activation functions, $\mathcal{N}(\cdot)$ is the neighbor set of node $i$ and $\mathbf{a}^{\top}$ represents learnable parameters. Attention coefficient $\alpha_{i,j}^{k}$ represents the weights between two nodes. During training, these weight coefficients are dynamically updated to reflect the importance of each node's contribution to its neighbors.

\begin{figure}[t]\centering
  \includegraphics[width=\linewidth]{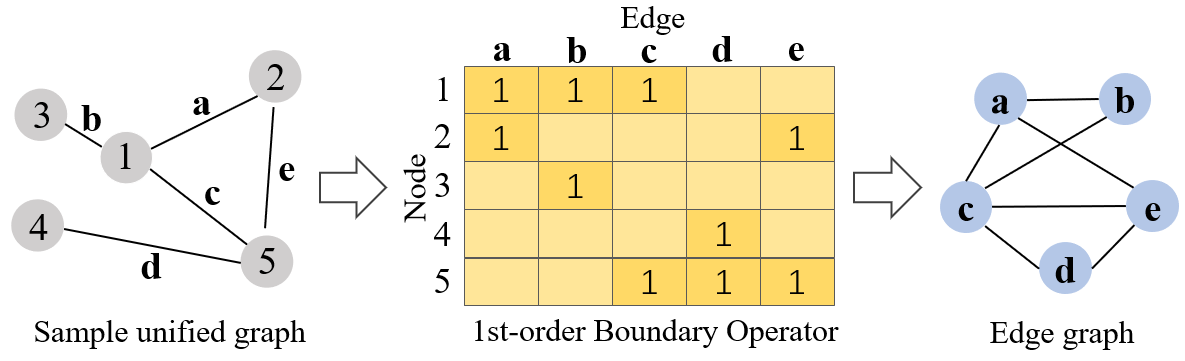}
  \caption{Illustration of edge graph construction from a unified ST graph using the first-order boundary operator $\mathcal{B}_{1}$. Nodes are represented by numbers, and edges connecting these nodes are labeled with letters. Applying the first-order boundary operator transforms each edge into a node in the edge graph, with connections formed based on shared nodes in the original graph.}
  \label{fig:edge_graph}
\end{figure}

\subsection{E2E-N2N Graph Convolution (E2E-N2N-GCN)}

Previous pedestrian trajectory models typically adopt node-centric approaches, such as N2N \cite{Mohamed2020socialstgcnn, shi2021sgcn, bae2022gpgraph, bae2023eigentrajectory, wu2023MSRL} and E2N \cite{Xu2022GroupNetMH, mo2021edge_mask} paradigms to understand and capture node dependencies. However, these methods overlook crucial E2E patterns, limiting their ability to capture the full spectrum of interaction dynamics. This oversight may result in a partial understanding of pedestrian behaviors, especially in complex scenarios where interaction patterns influence each other.

To address this limitation, we propose a novel Edge-to-Edge-Node-to-Node Graph Convolution (E2E-N2N-GCN) module (\figurename~\ref{fig:overview}~(b)), a dual-graph architecture that leverages the first-order boundary operator to construct edge graphs. By jointly modeling both explicit N2N social interactions among pedestrians and implicit E2E influence propagation across interaction patterns, our approach enables more comprehensive modeling of complex pedestrian behaviors. This dual-graph design allows each unified ST graph to capture how interaction patterns evolve and influence each other through connected edges, leading to more accurate trajectory predictions.

To construct the edge graph, we apply the first-order boundary operator $\mathcal{B}_{1}$ to transform it into its corresponding undirected edge graph $\mathcal{G}_{edge}^{k} = (\mathcal{E}^{k}, \mathcal{A}_{edge}^{k} )$, where $\mathcal{E}^{k}$ represents the node features in the edge graph, and $\mathcal{A}_{edge}^{k}$ indicates the new adjacency relations. This operator reinterprets the connections between nodes (edges in the original graph) as nodes in the new graph, creating edges between these new nodes if they share a common node in the original graph. \figurename~\ref{fig:edge_graph} illustrates this process, effectively showing how relationships are redefined to highlight deeper interaction dynamics.

To analyze and update the feature propagation of each edge graph, we employ the first-order Hodge Laplacian  \cite{xia2023deciphering, huang2023brainfunction} to analyze and learn the dynamics within these edge graphs:
\begin{equation}
\mathcal{L}_{1} = \mathcal{B}_{1}^{\top}\mathcal{B}_{1} + \mathcal{B}_{2}^{\top}\mathcal{B}_{2},
\end{equation}
where $\mathcal{L}_{1}$ represents first-order Hodge Laplacian operator, and $\mathcal{B}_{1}^\top$ captures and enhances edge relationships, focusing on direct interactions. $\mathcal{B}_{2}$ is typically relevant for higher-dimensional structures and not a primary focus here. We perform edge convolution by adapting the Hodge-Laplacian Laguerre Convolution (HLLConv) \cite{xia2023deciphering,huang2023brainfunction} to obtain the high-level edge embedding $\mathcal{H}_{edge}^{k}$ for each edge graph $k$:

\begin{equation}
\begin{aligned}
    \mathcal{H}_{edge}^{k} &= HLLConv(\mathcal{E}^{k}, \mathcal{A}_{edge}^{k}) \\
    &= \hbar_{1} \ast \mathcal{E}^{k} \\
    &= \sum_{j=0}^{J-1} \theta_{j} \Gamma_{j}(\mathcal{L}_{1}) \mathcal{E}^{k},
\end{aligned}
\end{equation}
where $\hbar_1$ is a spectral filter based on $\mathcal{L}_{1}$ applied to update edge features $\mathcal{E}^{k}$, with $\theta_j$ representing learnable parameters, and $\Gamma_j(\cdot)$ indicates the Laguerre polynomial functions. Detailed explanations of spectral filter $\hbar_1$ are shown in Algorithm~\ref{alg:hllconv}.

\begin{algorithm}[t]
\caption{Hodge-Laplacian Laguerre Convolution}
\label{alg:hllconv}
\begin{algorithmic}
\STATE \textbf{Input:} first-order Hodge Laplacian $\mathcal{L}_{1} = \mathcal{B}_{1}^{\top}\mathcal{B}_{1} + \mathcal{B}_{2}^{\top}\mathcal{B}_{2}$
\STATE \textbf{Output:} Spectral filter $\hbar_1$

\STATE \textbf{Step 1:} Perform eigen-decomposition on $\mathcal{L}_{1}$:
\begin{equation*}
    \mathcal{L}_{1}\phi_{1}^{i} = \lambda_{1}^{i}\phi_{1}^{i}
\end{equation*}
to obtain the orthonormal bases $\phi_{1}^{i}$ for $i \in [0, 1, 2, \cdots, \infty ]$. The spectral filter $\hbar$ of the 1-st order HL can be represented as $\hbar_{1}(\cdot, \cdot) = \sum_{i=0}^{\infty}\hbar_{1}(\lambda_{1}^{i})\phi_{1}^{i}(\cdot)\phi_{1}^{i}(\cdot)$.

\STATE \textbf{Step 2:} Approximate the spectral filter $\hbar_{1}(\lambda_{1})$ by Laguerre polynomial functions:
\begin{equation*}
    \hbar_{1}(\lambda_{1}) = \sum_{j=0}^{J-1}\theta_{j}\Gamma_{j}(\lambda_{1})
\end{equation*}
where $\theta_{j}$ is the $j$th expansion coefficient with $j$th Laguerre polynomial, and $\Gamma_{j}(\cdot)$ is written in a recurrence format as:
\begin{equation*}
    \Gamma_{j+1}(\lambda_{1}) = \frac{(2j+1 - \lambda_{1})\Gamma_{j}(\lambda_{1}) - j\Gamma_{j-1}(\lambda_{1})}{j+1}
\end{equation*}
where $\Gamma_{0}(\lambda_{1}) = 1$ and $\Gamma_{1}(\lambda_{1}) = 1 - \lambda_{1}$.
\end{algorithmic}
\end{algorithm}

Finally, after obtaining the embedded node features $\mathcal{H}_{node}^{k}$ and edge features $\mathcal{H}_{edge}^{k}$ for the $k$th unified ST graph, we leverage a fusion GCN to integrate node and edge embeddings, enhancing the understanding of graph dynamics. Specifically, we incorporate normalized edge embedding as weights into the aggregation process of GCN:
\begin{equation}
    \mathcal{H}^{k} = GCN(\mathcal{H}_{node}^{k}, \mathcal{H}_{edge}^{k}, \mathcal{A}_{node}^{k}),
\end{equation}
and each node $i$ in the graph is embedded as:
\begin{align}
    \mathcal{H}_{i}^{k} &= \sigma\left(\mathbf{\Theta}(\mathcal{H}_{node, i}^{k}) + \sum_{j \in \mathcal{N}(i)} \Phi(\mathcal{H}_{edge, ij}^{k}) \mathbf{\Theta}(\mathcal{H}_{node, j}^{k})\right),
\end{align}
where \(\mathbf{\Theta(\cdot)}\) and \(\Phi(\cdot)\) are linear transformations for node and edge features \cite{xia2023deciphering}, with \(\sigma(\cdot)\) as the activation function.

\subsection{Transformer Encoder Predictor}
\label{sec: trajectorypredictor}

Temporal dependency modeling in trajectory prediction has evolved through various architectures. RNNs \cite{Alexandre2016lstm, gupta2018socialgan} and TCNs \cite{Mohamed2020socialstgcnn, shi2021sgcn} have been widely adopted, they suffer from limited receptive fields and struggle to capture long-range dependencies. Although Transformer encoder-decoder architectures \cite{vaswani2017transformer, shi2023TUTR, chen2023ppnet} address the long-range dependency issue, it introduces extra computation costs.

In this work, we design a Transformer encoder-based predictor for trajectory prediction. As shown in \figurename~\ref{fig:overview}~(c), by encoding future trajectories as learnable parameters and concatenating them with historical trajectories, our approach enables unified modeling of both past and future information, allowing the model to fully leverage global temporal dependencies \cite{li2024autoregressiveImage} for more accurate predictions. We simply stack the graph embeddings $\mathcal{H}^{k}$ output by E2E-N2N-GCN across all patches to obtain the integrated feature representations $\textbf{H}$:
\begin{equation}
    \textbf{H} = \text{STACK}(\mathcal{H}^{1}, \mathcal{H}^{2}, \cdots, \mathcal{H}^{K}) \in \mathbb{R}^{K \times (NL) \times D}.
\end{equation}
We perform temporal average pooling across the $L$ channel,
and the output $\textbf{H} \in \mathbb{R}^{N \times K \times D}$ is served as the historical input
tokens. \RCC{We then initialize a learnable placeholder to form the padded future tokens as $\textbf{F} \in \mathbb{R}^{N \times T_{pred} \times D}$. The temporal channel of these tokens, $T_{pred}$, is tailored to match our prediction horizon. This setup aligns with the requirements of the Transformer encoder architecture \cite{vaswani2017transformer, liu2023itransformer}, which necessitates uniform sequence lengths for both inputs and outputs to enable synchronous processing. This design allows our model to directly produce trajectories of the required length. Throughout the training process, these placeholders are incrementally refined to represent the predicted trajectories, thereby enhancing the prediction capabilities.}

\begin{figure}[t]
  \centering
  \includegraphics[width=0.85\linewidth]{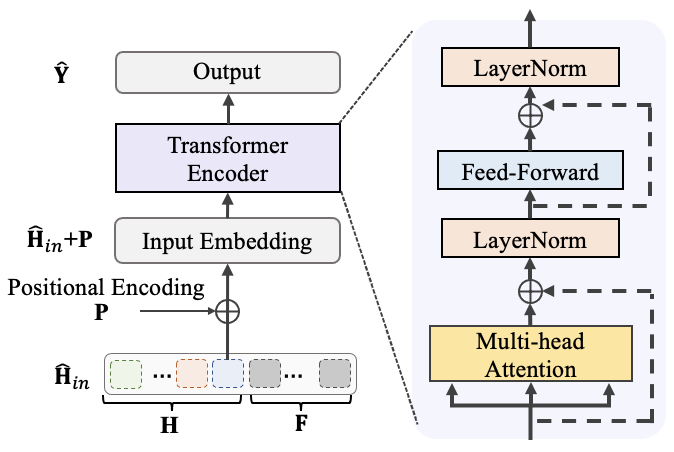}
  \caption{Illustration of the Transformer encoder-based predictor.}
  \label{fig:transformer_encoder}
\end{figure}

\RCC{Finally, the input tokens for the Transformer encoder are formed by concatenating the learned historical input tokens $\textbf{H}$ and padded future tokens $\textbf{F}$, resulting in the concatenated feature representation $\hat{\textbf{H}}_{in} \in \mathbb{R}^{N \times (K+T_{pred}) \times D}$. We further enhance these tokens with a learnable additive position embedding $\textbf{P} \in \mathbb{R}^{N \times (K+T_{pred}) \times D}$ \cite{vaswani2017transformer} that is applied to the entire concatenated sequence to preserve the temporal order information. The Transformer encoder then processes these augmented inputs to produce the predicted sequence representations $\hat{\textbf{Y}} \in \mathbb{R}^{N \times (K+T_{pred}) \times D}$:
\begin{gather}
\hat{\textbf{Y}} = \textit{Encoder}(\hat{\textbf{H}}_{in} + \textbf{P}), \nonumber \\
\hat{\textbf{H}}_{in} = [\textbf{H} \parallel \textbf{F}],
\end{gather}
where $[\cdot\parallel\cdot]$ denotes the concatenation operation along the temporal dimension. Note that $\hat{\textbf{Y}}$ represents the complete output of the encoder with length $K+T_{pred}$, only the last $T_{pred}$ time steps are used as the predicted trajectory representations, corresponding to the padded future tokens $\textbf{F}$.}
The architecture of the Transformer encoder and the learning process are shown in \figurename~\ref{fig:transformer_encoder}.
Similarly to \cite{shi2021sgcn, Mohamed2020socialstgcnn, ruochen2022multiclassSGCN}, we employ the bi-variate Gaussian loss function $\mathcal{L}_{prediction}$ to optimize the trajectory prediction:
\begin{equation}
    \mathcal{L}_{prediction} = - {\textstyle \sum_{t = 1}^{T_{pred}}}\log \mathcal{P}((x_{t}, y_{t})|\hat{\mu}_{t}, \hat{\sigma}_{t},\hat{\rho}_{t}),
\end{equation}
where $\hat{\mu}$ and $\hat{\sigma}$ are the mean and variance of bi-variate Gaussian distribution, and $\hat{\rho}$ represents the correlation coefficient. 

\subsection{Implementation Details}
The UniEdge framework, developed using PyTorch, is trained end-to-end on an NVIDIA TITAN XP GPU. We use a consistent batch size of 128 across all datasets, with initial learning rates set at 0.001 for the ETH/UCY datasets and 0.01 for the SDD datasets. The learning rate is adjusted every 50 epochs by a factor of 0.5. The AdamW optimizer is employed to train the model. The architecture for learning graph employs single-layer GAT, HLLConv, and GCN components. Node and edge embedding dimensions are set to 128. The Transformer encoder-based predictor is configured with a hidden dimension of 256 with 4 attention heads. 

\section{Experiments}
\label{sec:experiments}

\subsection{Experimental Setup}
\label{sec:setup}
We evaluate the proposed UniEdge on multiple benchmark datasets, including ETH \cite{pellegrini2009ETH}, UCY \cite{Lerner2007UCY}, and Stanford Drone Dataset (SDD) \cite{Robicquet2016SDD}. The ETH dataset contains two subsets (ETH and HOTEL) and the UCY dataset contains three subsets (UNIV, ZARA1, ZARA2), with the total number of pedestrians captured in these 5 subsets being 1,536. SDD is a benchmark dataset for pedestrian trajectories captured by a drone with a bird's eye viewing of university campus scenes and it contains 5,232 pedestrians across 8 different scenes.

We follow the experimental setup of \cite{shi2021sgcn,mohamed2022socialimplicit,Alexandre2016lstm}, using 3.2 seconds (8 frames) of observation trajectories to predict the next 4.8 seconds (12 frames). For ETH and UCY datasets, we follow existing works \cite{gupta2018socialgan,shi2023TUTR,shi2021sgcn,Mohamed2020socialstgcnn,bae2022gpgraph,bae2023eigentrajectory} and use the leave-one-out strategy for training and evaluation. For SDD, we follow the existing train-test split \cite{bae2023eigentrajectory, bae2022gpgraph, bae2023TERN} to train and test our proposed method. During training, we employ data augmentation following \cite{mohamed2022socialimplicit} to diversify and enrich our training datasets. This strategy is pivotal in enhancing the model's generalization capabilities.

During testing, we follow the standard protocol \cite{gupta2018socialgan,Alexandre2016lstm} and sampling strategy \cite{bae2022gpgraph} that generates 20 predictions from the predicted distributions; the best sample is used to compute the evaluation metrics. Average Displacement Error (ADE) and Final Displacement Error (FDE) \cite{Alexandre2016lstm,gupta2018socialgan,Mohamed2020socialstgcnn,shi2021sgcn} are used as evaluation metrics: 
\begin{equation}
\label{metrics}
\begin{split}
    &\text{ADE} = \frac{1}{N \times T_{pred}} \sum_{i=1}^{N} \sum_{t=1}^{T_{pred}} \sqrt{(x_{t}^{i} - \hat{x}_{t}^{i})^2 + (y_{t}^{i} - \hat{y}_{t}^{i})^2},  \\
    &\text{FDE} = \frac{1}{N} \sum_{i=1}^{N} \sqrt{(x_{T_{pred}}^{i} - \hat{x}_{T_{pred}}^{i})^2 + (y_{T_{pred}}^{i} - \hat{y}_{T_{pred}}^{i})^2},
\end{split}
\end{equation}
where $(\hat{x}_t^i, \hat{y}_t^i)$ and $(x_t^i, y_t^i)$ represent the predicted trajectory coordinates and ground-truth trajectory coordinate for the $i$-th pedestrian at time step $t$. 

\begin{table*}[t]
\caption{Results on The ETH (ETH, HOTEL) and UCY (UNIV, ZARA1, ZARA2) Datasets for Pedestrian Trajectory Prediction}
\centering
\renewcommand{\arraystretch}{1.12}
\begin{tabular}
{cc|cccccc}

\hline
\multirow{2}{*}{Method} & \multirow{2}{*}{Venue/Year} &\multicolumn{6}{c}{ADE($\downarrow$) / FDE($\downarrow$)} \\
&  & ETH & HOTEL & UNIV & ZARA1 & ZARA2 & \textbf{AVG} \\
\hline\hline

Social GAN \cite{gupta2018socialgan}               &CVPR'18  &0.87/1.62 &0.67/1.37 &0.76/1.52 &0.35/0.68 &0.42/0.84 &0.61/1.21 \\
Social-STGCNN \cite{Mohamed2020socialstgcnn}       &CVPR'20  &0.64/1.11 &0.49/0.85 &0.44/0.79 &0.34/0.53 &0.30/0.48 &0.44/0.75 \\
SGCN \cite{shi2021sgcn}                            &CVPR'21  &0.63/1.03 &0.32/0.55 &0.37/0.70 &0.29/0.53 &0.25/0.45 &0.37/0.65 \\
GP-Graph \cite{bae2022gpgraph}                     &ECCV'22  &0.43/0.63 &0.18/0.30 &0.24/0.42 &0.17/0.31 &0.15/0.29 &0.23/0.39 \\
Social-VAE \cite{pei2022socialVAE}                 &ECCV'22  &0.41/0.58 &0.13/0.19 &0.21/0.36 &0.17/0.29 &\underline{0.13}/0.22 &0.21/0.33      \\
MemoNet \cite{xu2022memonet}                       &CVPR'22  &0.40/0.61 & \textbf{0.11}/\textbf{0.17} &0.24/0.43 &0.18/0.32 &0.14/0.24 &0.21/0.35      \\
GroupNet \cite{Xu2022GroupNetMH}                   &CVPR'22  &0.46/0.73 &0.15/0.25 &0.26/0.49 &0.21/0.39 &0.17/0.33 &0.25/0.44      \\
Graph-TERN \cite{bae2023TERN}                      &AAAI'23  &0.42/0.58 &0.14/0.23 &0.26/0.45 &0.21/0.37 &0.17/0.29 &0.24/0.38      \\
MSRL \cite{wu2023MSRL}                             &AAAI'23    &\underline{0.28}/0.47 &0.14/0.22 &0.24/0.43 &0.17/0.30 &0.14/0.23 &\underline{0.19}/0.33    \\

LED \cite{mao2023leapfrog}                         &CVPR'23  &0.39/0.58 & \textbf{0.11}/\textbf{0.17} &0.26/0.43 &0.18/\underline{0.26} &\underline{0.13}/0.22 &0.21/0.33      \\
EqMotion \cite{xu2023eqmotion}                     &CVPR'23  &0.40/0.61 &\underline{0.12}/\underline{0.18} &0.23/0.43 &0.18/0.32 &\underline{0.13}/0.23 &0.21/0.35       \\
EigenTrajectory \cite{bae2023eigentrajectory}      &ICCV'23  &0.36/0.57 &0.13/0.21 &0.24/0.43 &0.20/0.35 &0.15/0.26 &0.22/0.36      \\
TUTR \cite{shi2023TUTR}                            &ICCV'23  &0.40/0.61 & \textbf{0.11}/\underline{0.18} &0.23/0.42 &0.18/0.34 &\underline{0.13}/0.25 &0.21/0.36 \\

SMEMO \cite{marchetti2024smemo} &TPAMI'24 &0.39/0.59 &0.14/0.20 &0.23/0.41 &0.19/0.32 &0.15/0.26 &0.22/0.35 \\

MFAN \cite{li2024mfan}                             &PR'24
&0.48/0.62	&0.17/0.21	&0.26/0.41	&0.23/0.36	&0.21/0.33	&0.27/0.39 \\

DDL \cite{wang2024pedestrianICRA}                  &ICRA'24
&\textbf{0.26}/0.50 &0.15/0.35 &0.29/0.58 &\underline{0.16}/0.29 &\underline{0.13}/0.22 &0.20/0.39 \\

\RCC{ATP-VAE} \RCC{\cite{pei2024autofocusing}} &\RCC{TCSVT'24}
&\RCC{0.48/0.76} &\RCC{0.14/0.20} &\RCC{0.26/0.44} &\RCC{0.28/0.48} &\RCC{0.20/0.35} 
&\RCC{0.27/0.45} \\

\RCC{MRGTraj} \RCC{\cite{peng2024mrgtraj}} &\RCC{TCSVT'24} 
&\RCC{\underline{0.28}/0.47} &\RCC{0.21/0.39} &\RCC{0.33/0.60} &\RCC{0.24/0.44} &\RCC{0.22/0.41} 
&\RCC{0.26/0.46} \\

SingularTrajectory \cite{bae2024singulartrajectory} &CVPR'24
&0.35/\textbf{0.42} &0.13/0.19 &0.25/0.44 &0.19/0.32 &0.15/0.25 &0.21/0.32 \\

HighGraph \cite{kim2024higher} &CVPR'24 &0.40/0.55 &0.13/\textbf{0.17} &\underline{0.20}/\underline{0.33} &0.17/0.27 &\textbf{0.11}/\underline{0.21} &0.20/\underline{0.30} \\

\hline
UniEdge (Ours) & - & 0.36/\underline{0.46} & \textbf{0.11}/\textbf{0.17} & \textbf{0.19}/\textbf{0.28} & \textbf{0.14}/\textbf{0.20} & \textbf{0.11}/\textbf{0.16} & \textbf{0.18}/\textbf{0.25} \\
\hline
\end{tabular}
\label{tab:ethanducy}
\end{table*}

\subsection{Baseline Methods}
We compare the proposed UniEdge framework with the following previous state-of-the-art methods:

\textbf{Graph-based methods}: Social-STGCNN \cite{Mohamed2020socialstgcnn}: an approach that models ST pedestrian interactions through graphs; SGCN \cite{Alexandre2016lstm}: an approach that models ST interactions through sparse directed spatial graph and sparse directed temporal graph; GP-Graph \cite{bae2022gpgraph}: an approach that considers group-based pedestrian behaviors; Graph-TERN \cite{bae2023TERN}: an approach that integrates multi-relational graph and control endpoint for trajectory prediction; EigenTrajectory(+SGCN) \cite{bae2023eigentrajectory}: a model that learns trajectories in eigenspaces and graph representations. MFAN \cite{li2024mfan}: an approach that models ST interactions for both edges and nodes. HighGraph \cite{kim2024higher}: a plug-and-play module that captures high-order dynamics of pedestrians - we use the HighGraph(+Social-VAE) variant for comparisons.

\textbf{Generative-based methods}: Social GAN \cite{gupta2018socialgan}: a method that uses pooling window module with Generative Adversarial Network (GAN) to generate diverse trajectories; Social-VAE \cite{pei2022socialVAE}: a method that employs timewise variational autoencoder(VAE) and attention mechanism to generate trajectories; GroupNet \cite{Xu2022GroupNetMH}: a method that introduces multiscale hypergraph with edge strength, utilizing conditional-VAE (CVAE) to generate trajectories; MSRL \cite{wu2023MSRL}: a method that models multi-stream interactions for trajectory prediction based on CVAE; MRGTraj \cite{peng2024mrgtraj}: a method based on CVAE and non-auto-regressive transformer encoder to generate diverse trajectories; ATP-VAE \cite{pei2024autofocusing}: an attention-based VAE architecture for trajectory prediction; LED \cite{mao2023leapfrog}: a multi-modal framework based on diffusion for prediction; SingularTrajectory \cite{bae2024singulartrajectory}: a diffusion framework based on singular projection and adaptive anchor to generate trajectories.

\textbf{Other methods}: MemoNet \cite{xu2022memonet}: an approach based on the retrospective-memory bank for trajectory representations; EqMotion \cite{xu2023eqmotion}: an approach that models trajectories via equivariant dynamics and invariant interaction; TUTR \cite{shi2023TUTR}: a transformer-based framework; SMEMO \cite{marchetti2024smemo}: an approach that models trajectories through social memory modules; DDL \cite{wang2024pedestrianICRA}: goal-based transformer for trajectory prediction.


\subsection{Quantitative Comparison}
\label{sec:quantitative}
\subsubsection{ETH and UCY Datasets}

\tablename~\ref{tab:ethanducy} presents the quantitative comparisons of our UniEdge model against existing methods under ADE and FDE metrics. Compared to the previous state-of-the-art (SOTA) generative-based method MSRL, our UniEdge demonstrates improvements of 5.3\% in average ADE and 24.2\% in average FDE. Unlike MSRL, which is a two-stage framework requiring separate training for the CVAE model and the trajectory decoder, UniEdge operates in an end-to-end manner, improving the overall performance while maintaining model parameter efficiency.
Compared to the best graph-based method HighGraph, our UniEdge shows significant improvements of 10.0\% in average ADE and 16.7\% in average FDE. Although HighGraph introduces high-order interaction modeling, it operates only on individual time steps, rather than cross-time interactions, which limits its effectiveness in capturing dynamic changes over time.
Contrasted to these graph-based methods, our UniEdge comprehensively models edge information flow and cross-time interactions, which can be the key to performance gain. Compared to DDL, which uses similar data pre-processing techniques, our UniEdge surpasses it by 10.0\% in ADE and 35.9\% in FDE, demonstrating enhanced prediction performance. While our UniEdge model demonstrates state-of-the-art (SOTA) performance on four subsets (HOTEL, UNIV, ZARA1, and ZARA2), particularly in environments with rich pedestrian interactions such as UNIV, it faces challenges similar to the graph-based SOTA method HighGraph on the ETH subset. This limitation of graph-based methods is mainly caused by the sparsity of the ETH subset, where fewer pedestrians and limited interactions constrain the expressive power of graph representations.

\begin{table}[t]
\caption{Results on The Stanford Drone Dataset (SDD) for Pedestrian Trajectory Prediction}
\centering
\footnotesize
\renewcommand{\arraystretch}{1.12}
\begin{tabular}{cc|c}
\hline
\multirow{2}{*}{Method} & \multirow{2}{*}{Venue/Year} & \multicolumn{1}{c}{ADE($\downarrow$) / FDE($\downarrow$)} \\
& & SDD \\
\hline\hline
Social GAN \cite{gupta2018socialgan}                    & CVPR'18 & 27.23/41.44 \\
Social-STGCNN \cite{Mohamed2020socialstgcnn}            & CVPR'20 & 26.46/42.71 \\
GroupNet \cite{Xu2022GroupNetMH}                        & CVPR'22 & 9.31/16.11 \\
MemoNet \cite{xu2022memonet}                            & CVPR'22 & 8.56/12.66 \\
GP-Graph\cite{bae2022gpgraph}                           & ECCV'22 & 9.10/13.80 \\
MSRL \cite{wu2023MSRL}                                  & AAAI'23 & 8.22/13.39 \\
Graph-TERN \cite{bae2023TERN}                           & AAAI'23 & 8.43/14.26 \\
LED \cite{mao2023leapfrog}                              & CVPR'23 & 8.48/11.66 \\
EigenTrajectory \cite{bae2023eigentrajectory}           & ICCV'23 & 8.05/13.25 \\
TUTR \cite{shi2023TUTR}                                 & ICCV'23 & \underline{7.76}/12.69 \\
SMEMO \cite{marchetti2024smemo}                         & TPAMI'24 & 8.11/13.06  \\
MFAN \cite{li2024mfan}                                  &
PR'24 &9.69/14.51  \\
HighGraph \cite{kim2024higher}                          & CVPR'24 & 7.98/\underline{11.42}    \\
\hline
UniEdge (Ours)                                          & -       & \textbf{7.51}/\textbf{10.89} \\
\hline
\end{tabular}
\label{tab:sdd}
\end{table}

\subsubsection{SDD Dataset} 
Table~\ref{tab:sdd} presents the quantitative comparison results of our model against various previous methods on SDD dataset. Unlike the ETH and UCY datasets, the SDD is a larger dataset featuring more complex pedestrian interactions. Compared to generative-based methods, UniEdge improves 8.6\% in ADE compared to MSRL and 6.6\% in FDE compared to LED. As a graph-based approach, our UniEdge outperforms the best graph-based HighGraph model by 5.9\% in ADE and 4.6\% in FDE. Compared to SOTA methods, UniEdge shows an improvement of 3.0\% in ADE over TUTR. These results further highlight the effectiveness of our proposed UniEdge model in handling complex social scenarios.

\begin{figure*}[t]
\centering
\includegraphics[width=16cm]{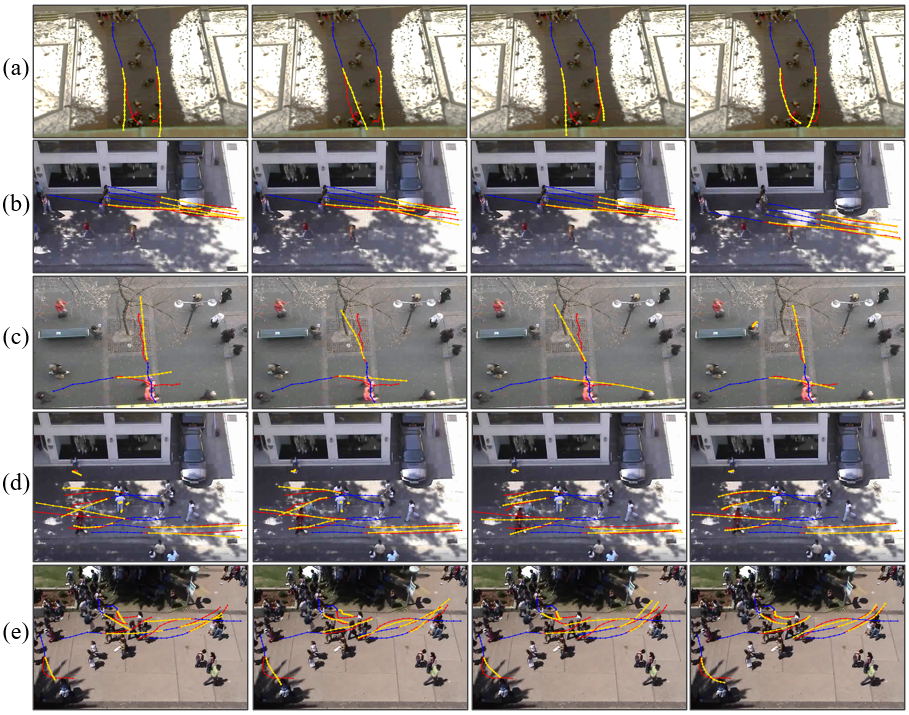}
\vspace{-5pt}
\hspace{0.3cm}\parbox{2.0cm}{\small GP-Graph \cite{bae2022gpgraph}} \hspace{1.0cm}
\parbox{3cm}{\centering\small Graph-TERN \cite{bae2023TERN}} \hspace{0.7cm}
\parbox{3cm}{\centering\small EigenTrajectory \cite{bae2023eigentrajectory}} \hspace{1.5cm}
\parbox{1.0cm}{\centering\small Ours}

\caption{\RCC{Visualization of predicted trajectories on the ETH and UCY datasets. Historical trajectories are in blue, ground-truth trajectories are in red, and predicted trajectories are in yellow. Scenario (a) shows two pedestrians walking in parallel and meet; Scenario (b) illustrates a group of pedestrians walking in parallel; (c) shows pedestrians meeting each other; (d) depicts several groups walking in opposing directions; and (e) presents a more complex scenario that pedestrian movements are stochastic.}}
\label{fig:eth-ucy_vis}
\end{figure*}

\subsection{Qualitative Comparison}
\label{sec:qualitative}

\subsubsection{Trajectory Visualization Comparison}
\RCC{In this section, we compare the most likely predictions between our UniEdge and previous graph-based methods, GP-Graph \cite{bae2022gpgraph}, Graph-TERN \cite{bae2023TERN} and EigenTrajectory \cite{bae2023eigentrajectory} on the ETH and UCY datasets.}

As shown in \figurename~\ref{fig:eth-ucy_vis}, our prediction results are significantly closer to the ground-truth trajectories compared to other methods in all scenarios. \RCC{\textbf{Scenario (a)} depicts two pedestrians walking and eventually meeting, where our predictions successfully capture their gradual convergence even in sparse environments. \textbf{Scenario (b)} shows pedestrians moving in parallel, where our approach achieves better alignment with ground-truth and avoids collisions compared to other methods. \textbf{Scenario (c)} presents two pedestrians meeting, where GP-Graph and EigenTrajectory fail to capture non-linear collision avoidance patterns. While Graph-TERN provides plausible predictions, our method better aligns with ground-truth by effectively modeling cross-time interactions. \textbf{Scenario (d)} presents a complex scenario in which several groups of pedestrians walk in opposing directions. In this case, GP-Graph and EigenTrajectory significantly suffer pedestrian collision issues. Our UniEdge demonstrates superior capability in capturing nonlinear movements, showcasing enhanced predictive accuracy in dynamically complex pedestrian interactions compared to previous methods. Finally, \textbf{scenario (e)} features complex non-linear trajectories with abrupt changes, where our method better captures overall movement trends despite shared challenges with certain trajectories.}

\begin{figure*}[t]
\centering
\includegraphics[width=16cm]{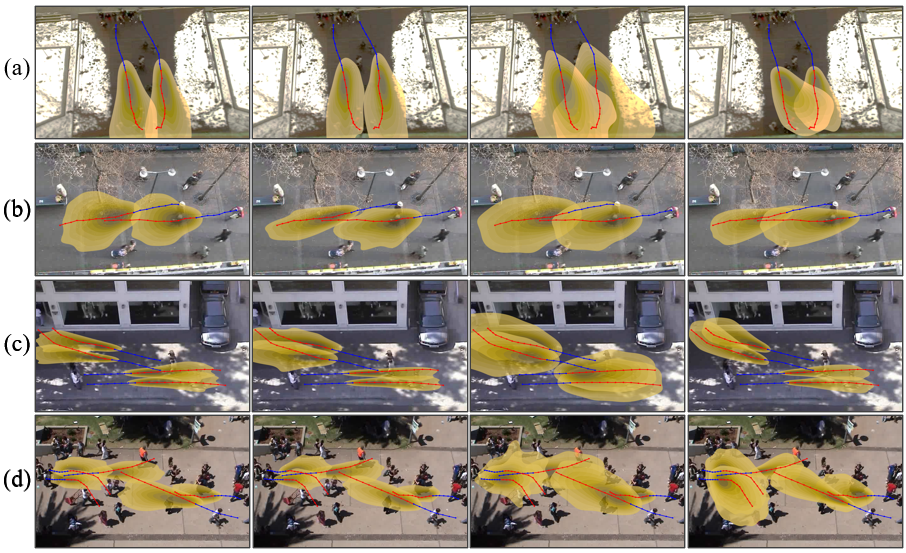}
\vspace{-5pt}
\hspace{0.2cm}\parbox{2.7cm}{\small GP-Graph \cite{bae2022gpgraph}} \hspace{0.2cm}
\parbox{3cm}{\centering \small Graph-TERN \cite{bae2023TERN}} \hspace{0.5cm}
\parbox{3cm}{\centering \small EigenTrajectory \cite{bae2023eigentrajectory}} \hspace{1.6cm}
\parbox{1.0cm}{\centering \small Ours}

\caption{\RCC{Visualization of predicted distributions on the ETH and UCY datasets. Historical trajectories are in blue, ground-truth trajectories are in red, and predicted trajectories are in yellow. Scenario (a) and (b) show two pedestrians walking in parallel with convergence; (c) presents two groups of pedestrians walking in opposing directions; (d) illustrates random walking behaviors.}}
\label{fig:eth-ucy_dist_vis}
\end{figure*}

\subsubsection{Distribution Visualization Comparisons}
\RCC{In this section, we further compare the predicted distributions of UniEdge with GP-Graph \cite{bae2022gpgraph}, Graph-TERN \cite{bae2023TERN} and EigenTrajectory \cite{bae2023eigentrajectory} on the ETH and UCY datasets. As shown in \figurename~\ref{fig:eth-ucy_dist_vis}, our method generates more accurate and plausible distributions. In \textbf{scenario (a)}, while other methods' distributions cover the ground-truth, they fail to capture the pedestrian convergence trend that our method successfully predicts. In \textbf{scenarios (b)} and \textbf{(c)}, GP-Graph and Graph-TERN generate either too narrow or broad distributions, failing to capture non-linear trajectories. EigenTrajectory covers ground-truth but produces overly broad, overlapping distributions that lead to collision issues. Our method achieves comprehensive coverage with fewer collision predictions. In \textbf{scenario (d)} with random walking patterns, our approach better captures both non-linear and linear trajectories.}

\subsection{Ablation Study and Model Analysis}
\label{sec:ablation}

\subsubsection{Model Component Analysis}
To verify the influence of each module incorporated in our UniEdge, we conduct ablation studies on the ETH and UCY datasets, which contain five different social scenarios. The results of these studies are detailed in \tablename~\ref{tab:ablation_onmodules}. In our experiments, variant (1) corresponds to the model excluding node-level embedding (NN), i.e., the model eliminates node-level GAT for capturing N2N interactions. variant (2) represents the model without edge-level embedding (EE), meaning that edge information is not integrated into the model's architecture, neglecting implicit edge feature propagation. Lastly, variant (3) describes the modeling process without learning edge graphs through Hodge-Laplacian Laguerre Convolution (HC). Specifically, node-level embedding provides an overall picture of pedestrians' interaction intentions to capture initial N2N interactions, the overall performance dropped 11.1\% in ADE and 24.0\% in FDE without N2N interactions. Variant (2) shows that without the modeling of implicit E2E influence propagation, the performance dropped 16.7\% in ADE and 20.0\% in FDE. Variant (3) demonstrate the effectiveness of the proposed edge-level reasoning, without Hodge-Laplacian Laguerre Convolutions, the overall performance dropped 16.7\% in ADE and 16.0\% in FDE, respectively. Notably, the UNIV subset, which contains the most pedestrians and the most complex interactions \cite{xu2022adaptiveTrans}, shows a decrease of 26.3\% in ADE and 35.7\% in FDE without edge graph learning, underscoring the importance of Hodge-Laplacian Laguerre convolution in managing the propagation of complex interactions. These findings underscore the importance of each module to the comprehensive functionality of our UniEdge model in trajectory prediction.

\begin{table}[t]
    \centering
    \setlength{\tabcolsep}{2pt} 
    \renewcommand{\arraystretch}{1.5} 
    \scriptsize
    \caption{Ablation Analysis of UniEdge on The ETH and UCY Datasets. NN = Node-level Embedding, EE = Edge-level Embedding, HC = Hodge-Laplacian Laguerre Convolution}
    \label{tab:ablation_onmodules}
    \begin{tabular}{cccc|cccccc}
        \hline
        \raisebox{-1.5ex}[0pt][0pt]{Variant} & \raisebox{-1.5ex}[0pt][0pt]{NN} & \raisebox{-1.5ex}[0pt][0pt]{EE} & \raisebox{-1.5ex}[0pt][0pt]{HC} & \multicolumn{6}{c}{ADE($\downarrow$) / FDE($\downarrow$)} \\
        & & & & ETH & HOTEL & UNIV & ZARA1 & ZARA2 & \multicolumn{1}{c}{AVG} \\
        \hline \hline
        (1) & $\times$ & \checkmark & \checkmark & 0.40/0.63  & 0.13/0.20 & \underline{0.22}/\underline{0.32} & \underline{0.15}/0.23 & \underline{0.12}/0.19 & \underline{0.20}/0.31 \\
        (2) & \checkmark & $\times$ & \checkmark & \underline{0.39}/0.54  & 0.14/\underline{0.18} & 0.23/0.35 & 0.16/0.24 & 0.13/0.19 & 0.21/0.30 \\
        (3) & \checkmark & \checkmark & $\times$ & \underline{0.39}/\underline{0.47}  & \underline{0.12}/\underline{0.18} & 0.24/0.38 & 0.17/\underline{0.22} & 0.14/\underline{0.18} & 0.21/\underline{0.29} \\
        
        Ours & \checkmark & \checkmark & \checkmark  & \textbf{0.36}/\textbf{0.46} & \textbf{0.11}/\textbf{0.17} & \textbf{0.19}/\textbf{0.28} & \textbf{0.14}/\textbf{0.20} & \textbf{0.11}/\textbf{0.16} & \textbf{0.18}/\textbf{0.25} \\
        \hline
    \end{tabular}
\end{table}

\RCC{To investigate the effectiveness of different node embedding approaches in our framework, we evaluate several graph neural networks as alternatives to our GAT-based N2N module, as shown in \tablename~\ref{tab:node_embedding}. The baseline GCN \cite{kipf2016GCN} exhibits limited performance due to its uniform neighborhood aggregation strategy. GraphSage \cite{hamilton2017graphsage} achieves improved results through its sampling-based aggregation strategy. Compared to GCN and GraphSage, GAT-based approach demonstrates superior performance through its attention mechanism, which enables dynamic weighting of pedestrian interactions while providing better interpretability through attention weights.}

\begin{table}
    \setlength{\tabcolsep}{2pt}
    \renewcommand{\arraystretch}{1.5}
    \centering  
    \scriptsize
    \caption{\RCC{Feature Embedding Analysis on The ETH and UCY Datasets}}
    \label{tab:node_embedding}
    \begin{tabular}{c|cccccc}
        \hline
        \raisebox{-1.5ex}[0pt][0pt]{Method} & \multicolumn{6}{c}{ADE($\downarrow$) / FDE($\downarrow$)} \\
        & ETH & HOTEL & UNIV & ZARA1 & ZARA2 & AVG \\ 
        \hline
        \hline
        w/ GCN \cite{kipf2016GCN} & 0.39/0.57 & 0.15/\underline{0.19} & 0.22/0.34 & \underline{0.17}/0.25 & 0.13/0.18 & 0.21/0.31 \\
        w/ GraphSage\cite{hamilton2017graphsage} & \underline{0.38}/\underline{0.52} & \underline{0.12}/\underline{0.19} & \underline{0.21}/\underline{0.30} & \textbf{0.14}/\underline{0.22} & \underline{0.12}/\underline{0.17} & \underline{0.19}/\underline{0.28} \\
        Ours & \textbf{0.36}/\textbf{0.44} & \textbf{0.11}/\textbf{0.17} & \textbf{0.19}/\textbf{0.28} & \textbf{0.14}/\textbf{0.20} & \textbf{0.11}/\textbf{0.16} & \textbf{0.18}/\textbf{0.25} \\
        \hline
    \end{tabular}
\end{table}

\subsubsection{Edge Feature Analysis}

To assess the impact of edge features in our UniEdge model, we conduct experiments focusing on their incorporation into edge graphs. As detailed in \tablename~\ref{tab:ablationedge}, we examine three edge feature types: a Gaussian kernel $\mathcal{E}_{i,j} = \exp\left(-\frac{d_{i,j}}{2\sigma^2}\right)$, which captures spatial relationships through the distance $d_{i,j}$ between nodes $i$ and $j$, and the standard deviation $\sigma$; a reciprocal distance kernel $\mathcal{E}_{i,j} = \frac{1}{d_{i,j} + \epsilon}$, highlighting inverse distance to represent pedestrian interactions; and a Euclidean distance kernel $\mathcal{E}_{i,j}=d_{i,j}$, quantifying node relationships based on direct distance. Results in \tablename~\ref{tab:ablationedge} show that the Euclidean distance (ours) kernel outperforms other features on the ETH and UCY datasets. We think this is because the Euclidean distance kernel directly and accurately measures distances between pedestrians, providing a more intuitive representation of pedestrian interactions.

\begin{table}
    \setlength{\tabcolsep}{2pt}
    \renewcommand{\arraystretch}{1.5} 
    \centering  
    \scriptsize
    \caption{Edge Feature Analysis on The ETH and UCY Datasets} 
    \label{tab:ablationedge} 
    \begin{tabular}{c|cccccc}
        \hline
        \raisebox{-1.5ex}[0pt][0pt]{Edge Feature} & \multicolumn{6}{c}{ADE($\downarrow$) / FDE($\downarrow$)} \\
        & ETH & HOTEL & UNIV & ZARA1 & ZARA2 & AVG \\ 
        \hline
        \hline
        Reciprocal distance & 0.40/0.55 & \underline{0.14}/0.21 & 0.21/0.31 & \underline{0.16}/\underline{0.23} & \underline{0.13}/0.20 & 0.21/0.30  \\
        Gaussian Kernel     & \underline{0.38}/\underline{0.52} & \underline{0.13}/\underline{0.19} & \underline{0.20}/\underline{0.30} & \underline{0.16}/\underline{0.23} & \underline{0.13}/\underline{0.19} & \underline{0.20}/\underline{0.29} \\
        
        Ours & \textbf{0.36}/\textbf{0.46} & \textbf{0.11}/\textbf{0.17} & \textbf{0.19}/\textbf{0.28} & \textbf{0.14}/\textbf{0.20} & \textbf{0.11}/\textbf{0.16}  & \textbf{0.18}/\textbf{0.25}\\
        \hline
    \end{tabular}
\end{table}

\begin{table}
    \setlength{\tabcolsep}{2pt}
    \renewcommand{\arraystretch}{1.5} 
    \centering  
    \scriptsize 
    \caption{Trajectory Predictor Analysis on The ETH and UCY Datasets. PE = Positional Encoding, Attn. Head = Attention Head, LN = Layer Normalization} 
    \label{tab:transformer_ablation} 
    \begin{tabular}{c|cccccc}
        \hline
        \raisebox{-1.5ex}[0pt][0pt]{Trajectory Predictor} & \multicolumn{6}{c}{ADE($\downarrow$) / FDE($\downarrow$)} \\
        & ETH & HOTEL & UNIV & ZARA1 & ZARA2 & AVG \\ 
        \hline
        \hline

        w/o PE  & 0.45/0.51 & 0.13/0.19 & 0.29/0.42 & 0.20/0.28 & 0.16/0.22 & 0.25/0.32 \\

        w/o Attn. Head  & \underline{0.37}/\underline{0.47} & \underline{0.12}/0.19 & 0.23/0.35 & 0.17/0.24 & \underline{0.13}/0.19 & 0.20/0.29 \\

        w/o LN       &0.38/\underline{0.47}    &0.13/\underline{0.18}   &\underline{0.21}/\underline{0.31}   &\underline{0.15}/\underline{0.23}       &\underline{0.13}/\underline{0.18}     &\underline{0.20}/\underline{0.27} \\

        Ours  & \textbf{0.36}/\textbf{0.44} & \textbf{0.11}/\textbf{0.17} & \textbf{0.19}/\textbf{0.28} & \textbf{0.14}/\textbf{0.20} & \textbf{0.11}/\textbf{0.16}  & \textbf{0.18}/\textbf{0.25}\\
        \hline
    \end{tabular}
\end{table}

\subsubsection{Trajectory Predictor Analysis}
To evaluate the effectiveness of the core modules in our Transformer encoder-based predictor and the corresponding padding approaches, we conduct extensive experiments on the predictor design. The results are presented in \tablename~\ref{tab:transformer_ablation}. We analyze three predictor variants: one without positional encoding (w/o PE), one without attention heads (w/o Attn. Head), and one without layer normalization (w/o LN). The experimental results demonstrate that the absence of any of these modules leads to degraded performance. Notably, the elimination of positional encoding has the most significant impact, resulting in performance degradation of 38.9\% in ADE and 28.0\% in FDE compared to the complete model. This substantial performance drop demonstrates the crucial role of positional encoding in preserving temporal ordering information of trajectory sequences, which is essential for understanding the temporal evolution of pedestrian motion patterns. Furthermore, the removal of attention heads leads to particularly inferior performance on the UNIV and ZARA1 subsets, which contain group activities with rich interactions, highlighting the importance of attention mechanisms in capturing temporal dependencies.

\RCC{To evaluate the performance on different predictor architectures, we conduct experiments on the ETH and UCY datasets, as shown in \tablename~\ref{tab:predictor_comparison}. The RNN-based \cite{HochSchm1997lstm} predictor shows limited performance due to its constrained receptive field and auto-regressive nature. The TCN-based predictor \cite{bai2018TCN} achieves strong performance on the ETH dataset due to its relatively large receptive field. However, its performance is limited on other datasets where temporal dependencies are more complex. Our Transformer Encoder-based predictor achieves superior performance by effectively capturing long-term dependencies through its non-local attention mechanism \cite{liu2023itransformer,vaswani2017transformer}.}

\begin{table}
    \setlength{\tabcolsep}{2pt}
    \renewcommand{\arraystretch}{1.5}
    \centering  
    \scriptsize
    \caption{\RCC{Trajectory Predictor Comparison Analysis on The ETH and UCY Datasets}}
    \label{tab:predictor_comparison}
    \begin{tabular}{c|cccccc}
        \hline
        \raisebox{-1.5ex}[0pt][0pt]{Trajectory Predictor} & \multicolumn{6}{c}{ADE($\downarrow$) / FDE($\downarrow$)} \\
        & ETH & HOTEL & UNIV & ZARA1 & ZARA2 & AVG \\ 
        \hline
        \hline
        RNN-based \cite{HochSchm1997lstm} & 0.84/1.18 & 0.18/0.30 & 0.40/0.66 & 0.62/1.13 & 0.24/0.41 & 0.46/0.74 \\
        TCN-based \cite{bai2018TCN} & \textbf{0.34}/\underline{0.48} & \underline{0.13}/\underline{0.19} & \underline{0.25}/\underline{0.35} & \underline{0.17}/\underline{0.26} & \underline{0.14}/\underline{0.19} & \underline{0.21}/\underline{0.29} \\
        Ours & \underline{0.36}/\textbf{0.44} & \textbf{0.11}/\textbf{0.17} & \textbf{0.19}/\textbf{0.28} & \textbf{0.14}/\textbf{0.20} & \textbf{0.11}/\textbf{0.16} & \textbf{0.18}/\textbf{0.25} \\
        \hline
    \end{tabular}
\end{table}

\subsubsection{\RCC{Unified ST Graph Analysis}}
\RCC{In this section, we analyze the effectiveness and impact of our proposed unified ST graph data structure while keeping other components fixed. The construction of this data structure is controlled by two key parameters: patch size $L$ and stride size $\mathcal{S}$. We conduct experiments on the ETH and UCY datasets to thoroughly analyze how these parameters affect the model's ability to capture ST inter-dependencies.} 

\begin{figure}[t]
  \centering
  \includegraphics[width=\linewidth]{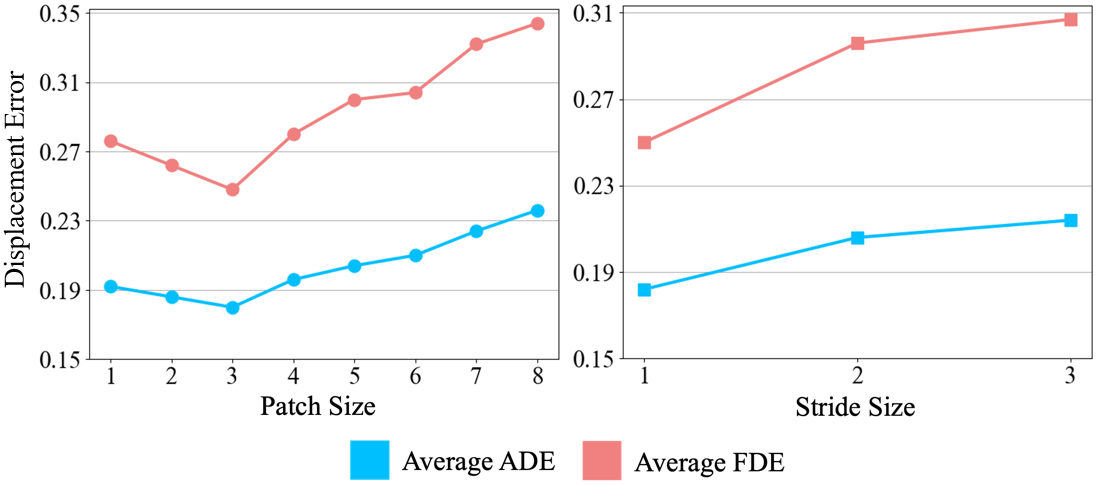}
  \caption{\RCC{Impact analysis of unified ST graph through patch size and stride size parameters on the ETH and UCY datasets.}}  
  \label{fig:patch_stride}
\end{figure}


\RCC{As shown in \figurename~\ref{fig:patch_stride} (\textbf{left}), we evaluate how patch size affects unified ST graph construction. A patch size of 1 reduces our model to traditional two-stage ST approaches \cite{shi2021sgcn,Mohamed2020socialstgcnn,huang2019stgat,bae2023eigentrajectory}, where cross-time interactions are not explicitly modeled. The model achieves optimal performance with a patch size of 3, effectively capturing local ST dependencies. Larger patch sizes, despite capturing more context information, may introduce redundant connections that degrade performance.}

\RCC{Second, we analyze the impact of stride size as shown in \figurename~\ref{fig:patch_stride} (\textbf{right}). The stride size determines the number of unified ST graphs and the overlap between adjacent patches. A larger stride size reduces the overlap between patches during graph construction, which in turn decreases the total number of unified ST graphs. A stride size of 1 yields the best performance in both ADE and FDE metrics, as it enables the capture of more fine-grained cross-time interactions through increased number of unified ST graphs. The increased number of unified ST graphs enables the transformer encoder-based predictor to leverage more ST contexts for enhanced performance.}

\subsubsection{Edge Weight Visualization}
\begin{figure}[t]
\centering
    \begin{overpic}[width=0.85\linewidth]{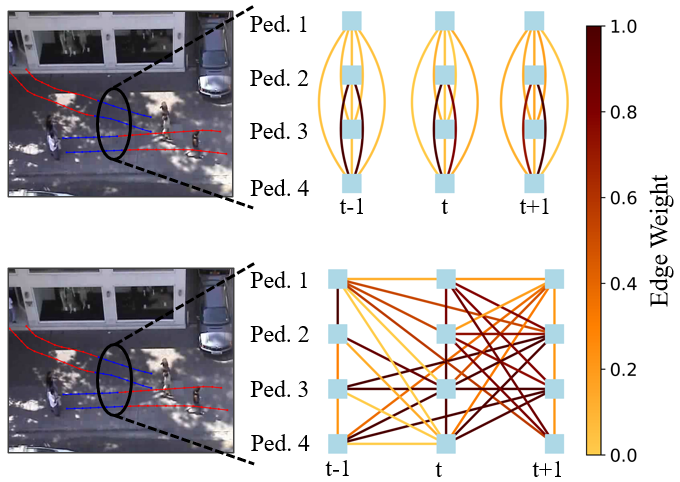}
        \put(30,35){\small EigenTrajectory\cite{bae2023eigentrajectory}}
        \put(40,-3){\small Ours}
    \end{overpic}
    \caption{Edge weight visualization of traditional two-stage ST approach EigenTrajectory and our UniEdge. Historical trajectories are in blue and ground-truth trajectories are in red.}
    \label{fig:edge_weight_vis}
\end{figure}

To provide qualitative insights into the differences between our UniEdge model and conventional ST architecture, we visualize the edge weights of our unified ST graph and EigenTrajectory \cite{bae2023eigentrajectory}. \figurename~\ref{fig:edge_weight_vis} illustrates a representative scenario where two groups of pedestrians approach each other across consecutive frames. While EigenTrajectory constructs independent spatial graphs for each frame, limiting its ability to capture high-order temporal dependencies, our unified ST graph architecture explicitly models cross-temporal interactions across all three frames. The visualization demonstrates how our model captures extended temporal dynamics, revealing interaction patterns that conventional ST frameworks may overlook.

\subsubsection{Predictor Attention Weight Visualization}

\begin{figure}[t]
\centering
    \includegraphics[width=0.9\linewidth]{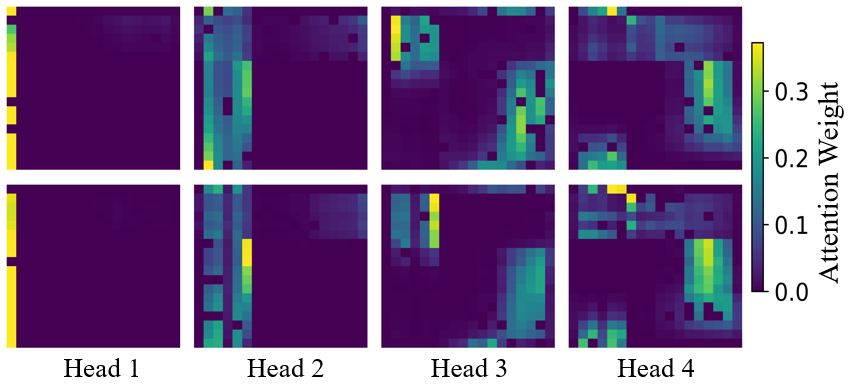}
    \caption{Predictor attention weight visualization. Four attention heads are configured in our experiments to analyze their impacts.}
    \label{fig:attn_weight_vis}
\end{figure}

This section visualizes the attention weights of our Transformer encoder-based predictor to examine interactions between learnable placeholder padding and historical contexts. As shown in \figurename~\ref{fig:attn_weight_vis}, the attention heads demonstrate distinct specialization patterns: heads 1 and 2 focus on temporal dependencies within historical trajectories, while heads 3 and 4 establish connections between learnable padding and relevant historical tokens. This specialized distribution reveals how the model decomposes trajectory prediction tasks and provides interpretable insights into its temporal information processing.

\subsubsection{\RCC{Complexity and Efficiency Analysis}}
\RCC{To evaluate the efficiency and computational complexity of UniEdge, \tablename~\ref{tab:complexity_eth} presents a comprehensive analysis of model complexity and computational efficiency among mainstream frameworks. We categorize the methods based on their temporal modeling paradigm into non-transformer and transformer-based temporal modeling methods. Compared to non-transformer temporal modeling methods such as EigenTrajectory \cite{bae2023eigentrajectory}, although UniEdge contains more parameters, it maintains competitive inference time while achieving significant improvements in prediction accuracy (18.2\% in ADE and 30.6\% in FDE). For common real-world trajectory prediction scenarios such as traffic collision avoidance and anomaly detection, we believe this trade-off is justified as prediction accuracy takes precedence over computational complexity, especially since higher accuracy in these applications can significantly reduce the risk of severe outcomes. Compared to transformer-based temporal modeling methods like TUTR \cite{shi2023TUTR} and MRGTraj \cite{peng2024mrgtraj}, UniEdge demonstrates superior efficiency with significantly lower parameters and FLOPs. Although TUTR achieves the fastest inference time, UniEdge maintains comparable computational speed while delivering substantially better prediction accuracy. Results demonstrate the effectiveness of our architecture in balancing computational efficiency and accuracy.}

\begin{table}[htpb]
\scriptsize
\begin{center}
\renewcommand{\arraystretch}{1.05} 
\caption{Complexity and Inference Time Analysis. All Models Are Evaluated on NVIDIA RTX3080 GPU}
\label{tab:complexity_eth}
\begin{tabular}{lcccc} 
\hline

\rule{0pt}{2.5ex}Methods & Param & FLOPs & Infer. Time & ADE($\downarrow$)/FDE($\downarrow$) \\
& $\times10^6$ & (M) & (ms) &  \\
\hline
\multicolumn{5}{c}{\cellcolor{gray!20}\textbf{Non-Transformer Temporal Modeling}} \\ 
\hline
Social-VAE \cite{pei2022socialVAE} & 2.15 & 292.95 & 40.27 & \textbf{0.21}/\textbf{0.33} \\
Graph-TERN \cite{bae2023TERN}      & \underline{0.05} & 22.59 & 40.15 & 0.24/0.38  \\
EqMotion \cite{xu2023eqmotion}     & 3.02 & \underline{7.75} & \underline{35.92} & \textbf{0.21}/\underline{0.35}  \\
EigenTrajectory \cite{bae2023eigentrajectory} & \textbf{0.02} & \textbf{1.36} & \textbf{22.26} & \underline{0.22}/0.36  \\
\hline
\multicolumn{5}{c}{\cellcolor{gray!20}\textbf{Transformer-based Temporal Modeling}} \\ 
\hline
TUTR \cite{shi2023TUTR}       & \underline{0.44} & \underline{64.54} & \textbf{20.21} & \underline{0.21}/\underline{0.36}  \\
\RCC{MRGTraj} \cite{peng2024mrgtraj} & \RCC{4.35} & \RCC{580.38} & \RCC{\underline{26.51}} & \RCC{0.26/0.46}  \\
UniEdge (Ours)                & \textbf{0.34} & \textbf{26.49} & 27.02 & \textbf{0.18}/\textbf{0.25}\\
\hline
\end{tabular}
\end{center}
\end{table}

\vspace{-11mm}
\RCC{\subsection{Discussion}}
\label{sec:discussion}

\RCC{In this section, we discuss potential reasons for the relatively lower performance of graph-based trajectory prediction approaches \cite{kim2024higher,bae2023eigentrajectory,bae2023TERN,li2024mfan} on the ETH subset, as compared to other scenarios. As indicated in \tablename~\ref{tab:dataset_stats}, the test set for the ETH subset averages only 2.59 pedestrians per sample, significantly less than other subsets, particularly the UNIV subset, which averages 25.70 pedestrians per sample. This stark variation in pedestrian density impacts the efficacy of graph-based methods, which rely on graph structures to model social interactions \cite{shi2021sgcn,mohamed2022socialimplicit}. The relatively sparse graph connectivity in the ETH scenario may impair message passing, potentially limiting the model's ability to effectively propagate and refine contextual information across nodes, which could hinder accurate representation of complex social interactions of graph-based approaches. In contrast, UniEdge demonstrates enhanced performance in scenarios with dense social interactions (HOTEL, UNIV, ZARA1, and ZARA2) by effectively capturing the more intricate social dynamics.}

\begin{figure}[t]
\centering
    \includegraphics[width=0.9\linewidth]{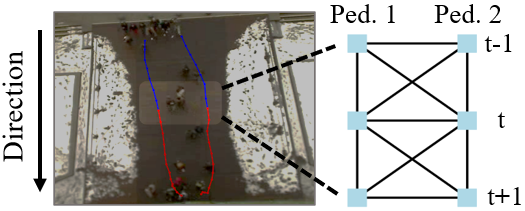}
    \caption{\RCC{Sample scenario in ETH dataset. Historical trajectories are in blue, ground-truth trajectories are in red.}}
    \label{fig:eth_scenario}
\end{figure}

\RCC{To further illustrate these challenges, we visualize a representative case from the ETH dataset in \figurename~\ref{fig:eth_scenario}. The example shows how UniEdge constructs a unified ST graph between Ped.1 and Ped.2, even though their trajectories are relatively stable with minimal interaction, potentially introducing unnecessary modeling bias. Additionally, while the scene contains multiple pedestrians, only a few trajectories are annotated, hindering the model's ability to capture comprehensive interaction patterns. To address these challenges, one promising direction is to develop dynamic graph optimization strategies \cite{ahmad2021skeletoneffective} that adapt connectivity based on scene characteristics. Such adaptive approaches would reduce redundant connections in sparse scenarios while preserving rich interaction modeling in dense scenarios, improving the prediction performance.}

\begin{table}[t]
\footnotesize
\begin{center}
\centering
\caption{\RCC{Dataset Statistics on The ETH and UCY Datasets}}
\label{tab:dataset_stats}
\begin{tabular}{c|ccccc}
\hline
Dataset & ETH & HOTEL & UNIV & ZARA1 & ZARA2 \\
\hline\hline
Total Test Samples & 70 & 301 & 947 & 602 & 921 \\
Avg. Pedestrians & 2.59 & 3.50 & 25.70 & 3.74 & 6.33 \\
\hline
\end{tabular}
\end{center}
\end{table}

\RCC{Additionally, we identify several promising directions to enhance our model's performance and adaptability. \textbf{First}, we aim to refine the model with an adaptive patch segmentation technique that dynamically adjusts patch sizes based on scene complexity metrics such as pedestrian density and interaction frequency \cite{huang2024hdmixer}, addressing the limitations of our current fixed patch size strategy and potentially improving prediction accuracy in varying crowd scenarios. \textbf{Second}, we plan to incorporate multimodal data sources, particularly environmental contextual images \cite{sun2021reciprocaltrajectory,Berenguer2021contextual}, to enhance our model's awareness of physical constraints and scene semantics, enabling more precise predictions in complex urban environments while reducing prediction errors caused by environmental factors. \textbf{Finally}, we will explore hardware optimization strategies for the transformer architecture \cite{dao2022flashattention,tay2022transeff} to improve deployment efficiency in real-time applications, reducing computation latency while maintaining prediction accuracy.}

\section{Conclusion}
In this paper, we introduce a novel UniEdge framework for trajectory prediction. Firstly, to capture high-order cross-time social interactions, we propose a patch-based unified ST graph architecture that simplifies high-order cross-time interactions to first-order relationships. Our approach reduces the steps required to aggregate spatial-temporal dependencies and effectively addresses the under-reaching problem by directly linking high-order nodes, offering a consistent improvement over traditional methods. Secondly, we propose the E2E-N2N Graph Convolution (E2E-N2N-GCN), a dual-graph architecture that jointly models explicit N2N social interactions and implicit E2E influence propagation through first-order boundary operators. This design enables comprehensive modeling of both individual behaviors and collective interaction dynamics. Finally, we propose a Transformer encoder-based trajectory predictor with placeholder-based techniques, providing a global view of trajectory embeddings, and improving the prediction performance. \RCC{Experiments on datasets demonstrate that UniEdge consistently outperforms state-of-the-art methods in both quantitative and qualitative evaluations.}

\section*{Acknowledgment}
{
This project is supported in part by the EPSRC NortHFutures project (ref: EP/X031012/1).
}

{\footnotesize
\bibliographystyle{IEEEtran}
\bibliography{main.bib}

\begin{thebibliography}{10}
\providecommand{\url}[1]{#1}
\csname url@samestyle\endcsname
\providecommand{\newblock}{\relax}
\providecommand{\bibinfo}[2]{#2}
\providecommand{\BIBentrySTDinterwordspacing}{\spaceskip=0pt\relax}
\providecommand{\BIBentryALTinterwordstretchfactor}{4}
\providecommand{\BIBentryALTinterwordspacing}{\spaceskip=\fontdimen2\font plus
\BIBentryALTinterwordstretchfactor\fontdimen3\font minus \fontdimen4\font\relax}
\providecommand{\BIBforeignlanguage}[2]{{%
\expandafter\ifx\csname l@#1\endcsname\relax
\typeout{** WARNING: IEEEtran.bst: No hyphenation pattern has been}%
\typeout{** loaded for the language `#1'. Using the pattern for}%
\typeout{** the default language instead.}%
\else
\language=\csname l@#1\endcsname
\fi
#2}}
\providecommand{\BIBdecl}{\relax}
\BIBdecl

\bibitem{bai2015pomdpintro}
H.~Bai, S.~Cai, N.~Ye, D.~Hsu, and W.~S. Lee, ``Intention-aware online pomdp planning for autonomous driving in a crowd,'' in \emph{IEEE Int. Conf. Robot. Autom.}\hskip 1em plus 0.5em minus 0.4em\relax IEEE, 2015, pp. 454--460.

\bibitem{chen2023ppnet}
W.~Chen, Z.~Yang, L.~Xue, J.~Duan, H.~Sun, and N.~Zheng, ``Multimodal pedestrian trajectory prediction using probabilistic proposal network,'' \emph{IEEE Trans. Circuits Syst. Video Technol.}, vol.~33, no.~6, pp. 2877--2891, 2023.

\bibitem{sun2021reciprocaltrajectory}
H.~Sun, Z.~Zhao, Z.~Yin, and Z.~He, ``Reciprocal twin networks for pedestrian motion learning and future path prediction,'' \emph{IEEE Trans. Circuits Syst. Video Technol.}, vol.~32, no.~3, pp. 1483--1497, 2021.

\bibitem{zhou2022realapp}
X.~Zhou, H.~Ren, T.~Zhang, X.~Mou, Y.~He, and C.-Y. Chan, ``Prediction of pedestrian crossing behavior based on surveillance video,'' \emph{Sensors}, 2022.

\bibitem{liu2020trajectorycnn}
X.~Liu, J.~Yin, J.~Liu, P.~Ding, J.~Liu, and H.~Liu, ``Trajectorycnn: a new spatio-temporal feature learning network for human motion prediction,'' \emph{IEEE Trans. Circuits Syst. Video Technol.}, vol.~31, no.~6, pp. 2133--2146, 2020.

\bibitem{wang2023stmotion}
N.~Wang, G.~Zhu, H.~Li, M.~Feng, X.~Zhao, L.~Ni, P.~Shen, L.~Mei, and L.~Zhang, ``Exploring spatio–temporal graph convolution for video-based human–object interaction recognition,'' \emph{IEEE Trans. Circuits Syst. Video Technol.}, vol.~33, no.~10, pp. 5814--5827, 2023.

\bibitem{shi2021sgcn}
L.~Shi, L.~Wang, C.~Long, S.~Zhou, M.~Zhou, Z.~Niu, and G.~Hua, ``Sgcn: Sparse graph convolution network for pedestrian trajectory prediction,'' in \emph{Proc. IEEE Conf. Comput. Vis. Pattern Recognit.}, 2021, pp. 8994--9003.

\bibitem{ruochen2022multiclassSGCN}
R.~Li, S.~Katsigiannis, and H.~P. Shum, ``Multiclass-sgcn: Sparse graph-based trajectory prediction with agent class embedding,'' in \emph{IEEE Int. Conf. Image Process.}\hskip 1em plus 0.5em minus 0.4em\relax IEEE, 2022, pp. 2346--2350.

\bibitem{kosaraju2019socialbigat}
V.~Kosaraju, A.~Sadeghian, R.~Mart{\'\i}n-Mart{\'\i}n, I.~Reid, H.~Rezatofighi, and S.~Savarese, ``Social-bigat: Multimodal trajectory forecasting using bicycle-gan and graph attention networks,'' \emph{Proc. Adv. Neu. Inf. Process. Syst.}, vol.~32, 2019.

\bibitem{huang2019stgat}
Y.~Huang, H.~Bi, Z.~Li, T.~Mao, and Z.~Wang, ``Stgat: Modeling spatial-temporal interactions for human trajectory prediction,'' in \emph{Proc. IEEE/CVF Int. Conf. Comput. Vis.}, 2019, pp. 6272--6281.

\bibitem{Mohamed2020socialstgcnn}
A.~Mohamed, K.~Qian, M.~Elhoseiny, and C.~Claudel, ``Social-stgcnn: A social spatio-temporal graph convolutional neural network for human trajectory prediction,'' in \emph{Proc. IEEE Conf. Comput. Vis. Pattern Recognit.}, 2020, pp. 14\,424--14\,432.

\bibitem{bae2022gpgraph}
I.~Bae, J.-H. Park, and H.-G. Jeon, ``Learning pedestrian group representations for multi-modal trajectory prediction,'' in \emph{Proc. Eur. Conf. Comput. Vis.}\hskip 1em plus 0.5em minus 0.4em\relax Springer, 2022, pp. 270--289.

\bibitem{bae2023eigentrajectory}
I.~Bae, J.~Oh, and H.-G. Jeon, ``Eigentrajectory: Low-rank descriptors for multi-modal trajectory forecasting,'' in \emph{Proc. IEEE/CVF Int. Conf. Comput. Vis.}, 2023.

\bibitem{bae2023TERN}
I.~Bae and H.-G. Jeon, ``A set of control points conditioned pedestrian trajectory prediction,'' in \emph{Proc. AAAI Conf. Artif. Intell.}, vol.~37, no.~5, 2023, pp. 6155--6165.

\bibitem{petar2017GAT}
P.~Veli{\v{c}}kovi{\'{c}}, G.~Cucurull, A.~Casanova, A.~Romero, P.~Li{\`{o}}, and Y.~Bengio, ``{Graph Attention Networks},'' \emph{Proc. Int. Conf. Learn. Represent.}, 2018.

\bibitem{HochSchm1997lstm}
S.~Hochreiter and J.~Schmidhuber, ``Long short-term memory,'' \emph{Neural computation}, vol.~9, no.~8, pp. 1735--1780, 1997.

\bibitem{kipf2016GCN}
T.~N. Kipf and M.~Welling, ``Semi-supervised classification with graph convolutional networks,'' in \emph{Proc. Int. Conf. Learn. Represent.}, 2017.

\bibitem{bai2018TCN}
S.~Bai, J.~Z. Kolter, and V.~Koltun, ``An empirical evaluation of generic convolutional and recurrent networks for sequence modeling,'' \emph{arXiv preprint arXiv:1803.01271}, 2018.

\bibitem{lu2024nodemixup}
W.~Lu, Z.~Guan, W.~Zhao, Y.~Yang, and L.~Jin, ``Nodemixup: Tackling under-reaching for graph neural networks,'' in \emph{Proc. AAAI Conf. Artif. Intell.}, vol.~38, no.~13, 2024, pp. 14\,175--14\,183.

\bibitem{black2023understandinger}
M.~Black, Z.~Wan, A.~Nayyeri, and Y.~Wang, ``Understanding oversquashing in gnns through the lens of effective resistance,'' in \emph{Proc. Int. Conf. Mach. Learn.}\hskip 1em plus 0.5em minus 0.4em\relax PMLR, 2023, pp. 2528--2547.

\bibitem{Wang2023FullyConnectedSG}
Y.~Wang, Y.~Xu, J.~Yang, M.~Wu, X.~Li, L.~Xie, and Z.~Chen, ``Fully-connected spatial-temporal graph for multivariate time-series data,'' in \emph{Proc. AAAI Conf. Artif. Intell.}, vol.~38, no.~14, 2024, pp. 15\,715--15\,724.

\bibitem{yi2023fouriergnn}
K.~Yi, Q.~Zhang, W.~Fan, H.~He, L.~Hu, P.~Wang, N.~An, L.~Cao, and Z.~Niu, ``Fourier{GNN}: Rethinking multivariate time series forecasting from a pure graph perspective,'' in \emph{Proc. Adv. Neu. Inf. Process. Syst.}, 2023.

\bibitem{Xu2022GroupNetMH}
C.~Xu, M.~Li, Z.~Ni, Y.~Zhang, and S.~Chen, ``Groupnet: Multiscale hypergraph neural networks for trajectory prediction with relational reasoning,'' \emph{Proc. IEEE/CVF Conf. Comput. Vis. Pattern Recognit.}, pp. 6488--6497, 2022.

\bibitem{mo2021edge_mask}
X.~Mo, Y.~Xing, and C.~Lv, ``Heterogeneous edge-enhanced graph attention network for multi-agent trajectory prediction,'' \emph{arXiv preprint arXiv:2106.07161}, 2021.

\bibitem{xia2023deciphering}
Y.~Xia, Y.~Liang, H.~Wen, X.~Liu, K.~Wang, Z.~Zhou, and R.~Zimmermann, ``Deciphering spatio-temporal graph forecasting: A causal lens and treatment,'' in \emph{Proc. Adv. Neu. Inf. Process. Syst.}, 2023.

\bibitem{huang2023brainfunction}
J.~Huang, M.~K. Chung, and A.~Qiu, ``Heterogeneous graph convolutional neural network via hodge-laplacian for brain functional data,'' in \emph{Int. Conf. Inf. Process. Med. Imaging}.\hskip 1em plus 0.5em minus 0.4em\relax Springer, 2023, pp. 278--290.

\bibitem{wu2024deepdual}
X.~Wu, W.~Lu, Y.~Quan, Q.~Miao, and P.~G. Sun, ``Deep dual graph attention auto-encoder for community detection,'' \emph{Expert Syst. Appl.}, vol. 238, p. 122182, 2024.

\bibitem{post2007firstboundary}
O.~Post, ``First-order operators and boundary triples,'' \emph{Russian Journal of Mathematical Physics}, vol.~14, no.~4, pp. 482--492, 2007.

\bibitem{vaswani2017transformer}
A.~Vaswani, N.~Shazeer, N.~Parmar, J.~Uszkoreit, L.~Jones, A.~N. Gomez, {\L}.~Kaiser, and I.~Polosukhin, ``Attention is all you need,'' \emph{Advances in neural information processing systems}, vol.~30, 2017.

\bibitem{pellegrini2009ETH}
S.~Pellegrini, A.~Ess, K.~Schindler, and L.~Van~Gool, ``You'll never walk alone: Modeling social behavior for multi-target tracking,'' in \emph{Proc. IEEE/CVF Int. Conf. Comput. Vis.}, 2009, pp. 261--268.

\bibitem{Lerner2007UCY}
A.~Lerner, Y.~Chrysanthou, and D.~Lischinski, ``Crowds by example,'' in \emph{Computer graphics forum}, vol.~26, no.~3.\hskip 1em plus 0.5em minus 0.4em\relax Wiley Online Library, 2007, pp. 655--664.

\bibitem{Robicquet2016SDD}
A.~Robicquet, A.~Sadeghian, A.~Alahi, and S.~Savarese, ``Learning social etiquette: Human trajectory understanding in crowded scenes,'' in \emph{Proc. Eur. Conf. Comput. Vis.}\hskip 1em plus 0.5em minus 0.4em\relax Springer, 2016, pp. 549--565.

\bibitem{Alexandre2016lstm}
A.~Alahi, K.~Goel, V.~Ramanathan, A.~Robicquet, L.~Fei-Fei, and S.~Savarese, ``Social lstm: Human trajectory prediction in crowded spaces,'' in \emph{Proc. IEEE Conf. Comput. Vis. Pattern Recognit.}, 2016, pp. 961--971.

\bibitem{gupta2018socialgan}
A.~Gupta, J.~Johnson, L.~Fei-Fei, S.~Savarese, and A.~Alahi, ``Social gan: Socially acceptable trajectories with generative adversarial networks,'' in \emph{Proc. IEEE Conf. Comput. Vis. Pattern Recognit.}, 2018, pp. 2255--2264.

\bibitem{qiao20222ggcn}
T.~Qiao, Q.~Men, F.~W.~B. Li, Y.~Kubotani, S.~Morishima, and H.~P.~H. Shum, ``Geometric features informed multi-person human-object interaction recognition in videos,'' in \emph{Proc. Eur. Conf. Comput. Vis.}, 2022.

\bibitem{yan2018spatial}
S.~Yan, Y.~Xiong, and D.~Lin, ``Spatial temporal graph convolutional networks for skeleton-based action recognition,'' in \emph{Proceedings of the AAAI conference on artificial intelligence}, vol.~32, no.~1, 2018.

\bibitem{liu2023skeletonrecognition}
Y.~Liu, H.~Zhang, Y.~Li, K.~He, and D.~Xu, ``Skeleton-based human action recognition via large-kernel attention graph convolutional network,'' \emph{IEEE Trans. Vis. Comput. Graph.}, vol.~29, no.~5, pp. 2575--2585, 2023.

\bibitem{li2022graphdrug2}
X.-S. Li, X.~Liu, L.~Lu, X.-S. Hua, Y.~Chi, and K.~Xia, ``Multiphysical graph neural network (mp-gnn) for covid-19 drug design,'' \emph{Briefings in bioinformatics}, vol.~23, no.~4, p. bbac231, 2022.

\bibitem{wang2019rs1}
X.~Wang, X.~He, M.~Wang, F.~Feng, and T.-S. Chua, ``Neural graph collaborative filtering,'' in \emph{Proc.Int. ACM SIGIR Conf. Res. Dev. Inf. Retrieval}, 2019, pp. 165--174.

\bibitem{yu2018spatiotraffic}
B.~Yu, H.~Yin, and Z.~Zhu, ``Spatio-temporal graph convolutional networks: A deep learning framework for traffic forecasting,'' in \emph{Int. Joint Conf. Artif. Intell.}, 2018.

\bibitem{shi2023TUTR}
L.~Shi, L.~Wang, S.~Zhou, and G.~Hua, ``Trajectory unified transformer for pedestrian trajectory prediction,'' in \emph{Proc. IEEE/CVF Int. Conf. Comput. Vis.}, 2023, pp. 9675--9684.

\bibitem{xu2023gcvrnn}
Y.~Xu, A.~Bazarjani, H.-g. Chi, C.~Choi, and Y.~Fu, ``Uncovering the missing pattern: Unified framework towards trajectory imputation and prediction,'' in \emph{Proc. IEEE Conf. Comput. Vis. Pattern Recognit.}, 2023, pp. 9632--9643.

\bibitem{li2024mfan}
J.~Li, L.~Yang, Y.~Chen, and Y.~Jin, ``Mfan: Mixing feature attention network for trajectory prediction,'' \emph{Pattern Recognition}, vol. 146, p. 109997, 2024.

\bibitem{pei2022socialVAE}
P.~Xu, J.-B. Hayet, and I.~Karamouzas, ``Socialvae: Human trajectory prediction using timewise latents,'' in \emph{Proc. Eur. Conf. Comput. Vis.}, 2022, pp. 511--528.

\bibitem{pei2024autofocusing}
Z.~Pei, J.~Zhang, W.~Zhang, M.~Wang, J.~Wang, and Y.-H. Yang, ``Autofocusing for synthetic aperture imaging based on pedestrian trajectory prediction,'' \emph{IEEE Trans. Circuits Syst. Video Technol.}, vol.~34, no.~5, pp. 3551--3562, 2024.

\bibitem{Berenguer2021contextual}
A.~Díaz~Berenguer, M.~Alioscha-Perez, M.~C. Oveneke, and H.~Sahli, ``Context-aware human trajectories prediction via latent variational model,'' \emph{IEEE Trans. Circuits Syst. Video Technol}, vol.~31, no.~5, pp. 1876--1889, 2021.

\bibitem{peng2024mrgtraj}
Y.~Peng, G.~Zhang, J.~Shi, X.~Li, and L.~Zheng, ``Mrgtraj: A novel non-autoregressive approach for human trajectory prediction,'' \emph{IEEE Trans. Circuits Syst. Video Technol.}, vol.~34, no.~4, pp. 2318--2331, 2024.

\bibitem{wang2024pedestrianICRA}
H.~Wang, W.~Zhi, G.~Batista, and R.~Chandra, ``Pedestrian trajectory prediction using dynamics-based deep learning,'' in \emph{IEEE Int. Conf. Robot. Autom.}\hskip 1em plus 0.5em minus 0.4em\relax IEEE, 2024, pp. 15\,068--15\,075.

\bibitem{ghosh2008minimizing}
A.~Ghosh, S.~Boyd, and A.~Saberi, ``Minimizing effective resistance of a graph,'' \emph{SIAM review}, vol.~50, no.~1, pp. 37--66, 2008.

\bibitem{bozzo2013moore}
E.~Bozzo, ``The moore--penrose inverse of the normalized graph laplacian,'' \emph{Linear Algebra Appl.}, vol. 439, no.~10, pp. 3038--3043, 2013.

\bibitem{krizhevsky2012imagenet}
A.~Krizhevsky, I.~Sutskever, and G.~E. Hinton, ``Imagenet classification with deep convolutional neural networks,'' \emph{Adv. Neural Inf. Process. Syst.}, vol.~25, 2012.

\bibitem{brody2021gatv2}
S.~Brody, U.~Alon, and E.~Yahav, ``How attentive are graph attention networks?'' \emph{arXiv preprint arXiv:2105.14491}, 2021.

\bibitem{wu2023MSRL}
Y.~Wu, L.~Wang, S.~Zhou, J.~Duan, G.~Hua, and W.~Tang, ``Multi-stream representation learning for pedestrian trajectory prediction,'' in \emph{Proc. AAAI Conf. Artif. Intell.}, vol.~37, no.~3, 2023, pp. 2875--2882.

\bibitem{li2024autoregressiveImage}
T.~Li, Y.~Tian, H.~Li, M.~Deng, and K.~He, ``Autoregressive image generation without vector quantization,'' \emph{arXiv preprint arXiv:2406.11838}, 2024.

\bibitem{liu2023itransformer}
Y.~Liu, T.~Hu, H.~Zhang, H.~Wu, S.~Wang, L.~Ma, and M.~Long, ``itransformer: Inverted transformers are effective for time series forecasting,'' \emph{arXiv preprint arXiv:2310.06625}, 2023.

\bibitem{mohamed2022socialimplicit}
A.~Mohamed, D.~Zhu, W.~Vu, M.~Elhoseiny, and C.~Claudel, ``Social-implicit: Rethinking trajectory prediction evaluation and the effectiveness of implicit maximum likelihood estimation,'' in \emph{Proc. Eur. Conf. Comput. Vis.}\hskip 1em plus 0.5em minus 0.4em\relax Springer, 2022, pp. 463--479.

\bibitem{xu2022memonet}
C.~Xu, W.~Mao, W.~Zhang, and S.~Chen, ``Remember intentions: Retrospective-memory-based trajectory prediction,'' in \emph{Proc. IEEE Conf. Comput. Vis. Pattern Recognit.}, 2022, pp. 6488--6497.

\bibitem{mao2023leapfrog}
W.~Mao, C.~Xu, Q.~Zhu, S.~Chen, and Y.~Wang, ``Leapfrog diffusion model for stochastic trajectory prediction,'' in \emph{Proc. IEEE Conf. Comput. Vis. Pattern Recognit.}, 2023, pp. 5517--5526.

\bibitem{xu2023eqmotion}
C.~Xu, R.~T. Tan, Y.~Tan, S.~Chen, Y.~G. Wang, X.~Wang, and Y.~Wang, ``Eqmotion: Equivariant multi-agent motion prediction with invariant interaction reasoning,'' in \emph{Proc. IEEE Conf. Comput. Vis. Pattern Recognit.}, 2023, pp. 1410--1420.

\bibitem{marchetti2024smemo}
F.~Marchetti, F.~Becattini, L.~Seidenari, and A.~Del~Bimbo, ``Smemo: social memory for trajectory forecasting,'' \emph{IEEE Trans. Pattern Anal. Mach. Intell.}, 2024.

\bibitem{bae2024singulartrajectory}
I.~Bae, Y.-J. Park, and H.-G. Jeon, ``Singulartrajectory: Universal trajectory predictor using diffusion model,'' in \emph{Proc. IEEE Conf. Comput. Vis. Pattern Recognit.}, 2024, pp. 17\,890--17\,901.

\bibitem{kim2024higher}
S.~Kim, H.-g. Chi, H.~Lim, K.~Ramani, J.~Kim, and S.~Kim, ``Higher-order relational reasoning for pedestrian trajectory prediction,'' in \emph{Proc. IEEE Conf. Comput. Vis. Pattern Recognit.}, 2024, pp. 15\,251--15\,260.

\bibitem{xu2022adaptiveTrans}
Y.~Xu, L.~Wang, Y.~Wang, and Y.~Fu, ``Adaptive trajectory prediction via transferable gnn,'' in \emph{Proc. IEEE Conf. Comput. Vis. Pattern Recognit.}, 2022, pp. 6520--6531.

\bibitem{hamilton2017graphsage}
W.~Hamilton, Z.~Ying, and J.~Leskovec, ``Inductive representation learning on large graphs,'' \emph{Adv. Neural Inf. Process. Syst.}, vol.~30, 2017.

\bibitem{ahmad2021skeletoneffective}
T.~Ahmad, L.~Jin, L.~Lin, and G.~Tang, ``Skeleton-based action recognition using sparse spatio-temporal gcn with edge effective resistance,'' \emph{Neurocomputing}, vol. 423, pp. 389--398, 2021.

\bibitem{huang2024hdmixer}
Q.~Huang, L.~Shen, R.~Zhang, J.~Cheng, S.~Ding, Z.~Zhou, and Y.~Wang, ``Hdmixer: Hierarchical dependency with extendable patch for multivariate time series forecasting,'' in \emph{Proc. AAAI Conf. Artif. Intell.}, vol.~38, no.~11, 2024, pp. 12\,608--12\,616.

\bibitem{dao2022flashattention}
T.~Dao, D.~Fu, S.~Ermon, A.~Rudra, and C.~R{\'e}, ``Flashattention: Fast and memory-efficient exact attention with io-awareness,'' \emph{Adv. Neural Inf. Process. Syst.}, vol.~35, pp. 16\,344--16\,359, 2022.

\bibitem{tay2022transeff}
Y.~Tay, M.~Dehghani, D.~Bahri, and D.~Metzler, ``Efficient transformers: A survey,'' \emph{ACM Comput. Surv.}, vol.~55, no.~6, 2022.

\end{thebibliography}
}
\vfill

\begin{IEEEbiography}
[{\includegraphics[width=1in,height=1.3in,clip,keepaspectratio]{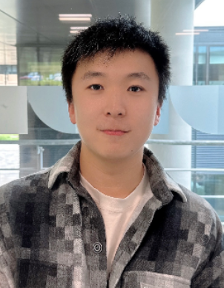}}]{Ruochen Li}
is a PhD candidate in the Department of Computer Science at Durham University, UK. He received the BSc and MSc in Computer Science at the University of Leeds, UK. His research interests include trajectory prediction, computer vision, time-series analysis, and graph convolutional networks.
\end{IEEEbiography}
\vspace{-11mm}
\begin{IEEEbiography}
[{\includegraphics[width=1in,height=1.3in,clip,keepaspectratio]{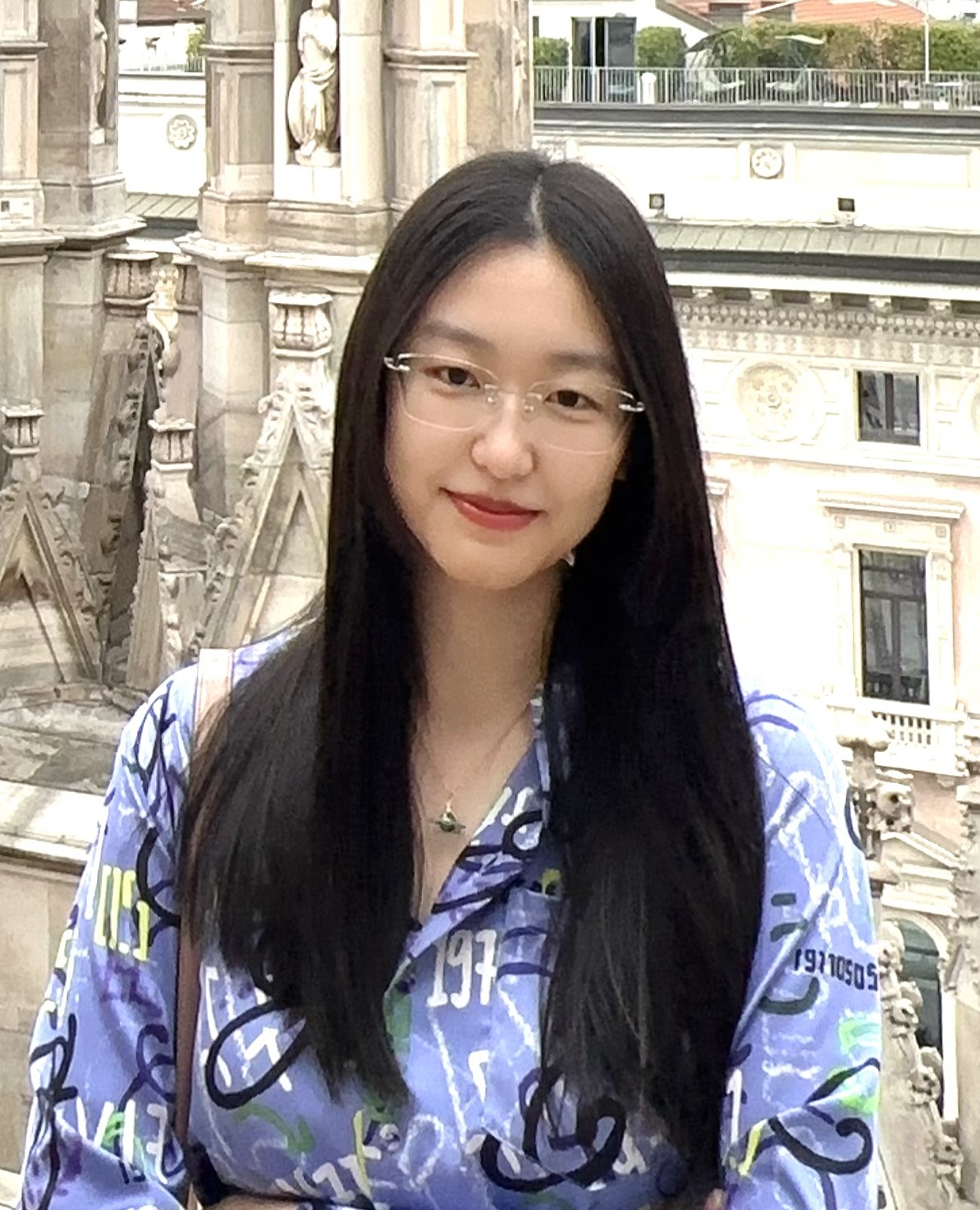}}]{Tanqiu Qiao} is currently pursuing her Ph.D. degree in the Department of Computer Science at Durham University, UK. She received the BSc and MSc in Statistics at the University of Glasgow and the University of Sheffield, UK in 2019 and 2020, respectively. Her main research interests include human-object interaction and human motion recognition. 
\end{IEEEbiography}
\vspace{-11mm}
\begin{IEEEbiography}
[{\includegraphics[width=1in,height=1.3in,clip,keepaspectratio]{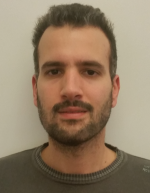}}]{Stamos Katsigiannis} (Member, IEEE) received the B.Sc. (Hons.) degree in informatics and telecommunications from the National and Kapodistrian University of Athens, Greece, in 2009,  the M.Sc. degree in computer science from the Athens University of Economics and Business, Greece, in 2011, and the Ph.D. degree in computer science (biomedical image and general-purpose video processing) from the National and Kapodistrian University of Athens, Greece, in 2016. He is currently an Associate Professor with the Department of Computer Science, Durham University, UK. He has participated in multiple national and international research projects and has authored and co-authored over 60 research publications. His research interests include machine learning, natural language processing, affective computing, image analysis, and image and video quality.
\end{IEEEbiography}
\vspace{-11mm}
\begin{IEEEbiography}
[{\includegraphics[width=1in,height=1.3in,clip,keepaspectratio]{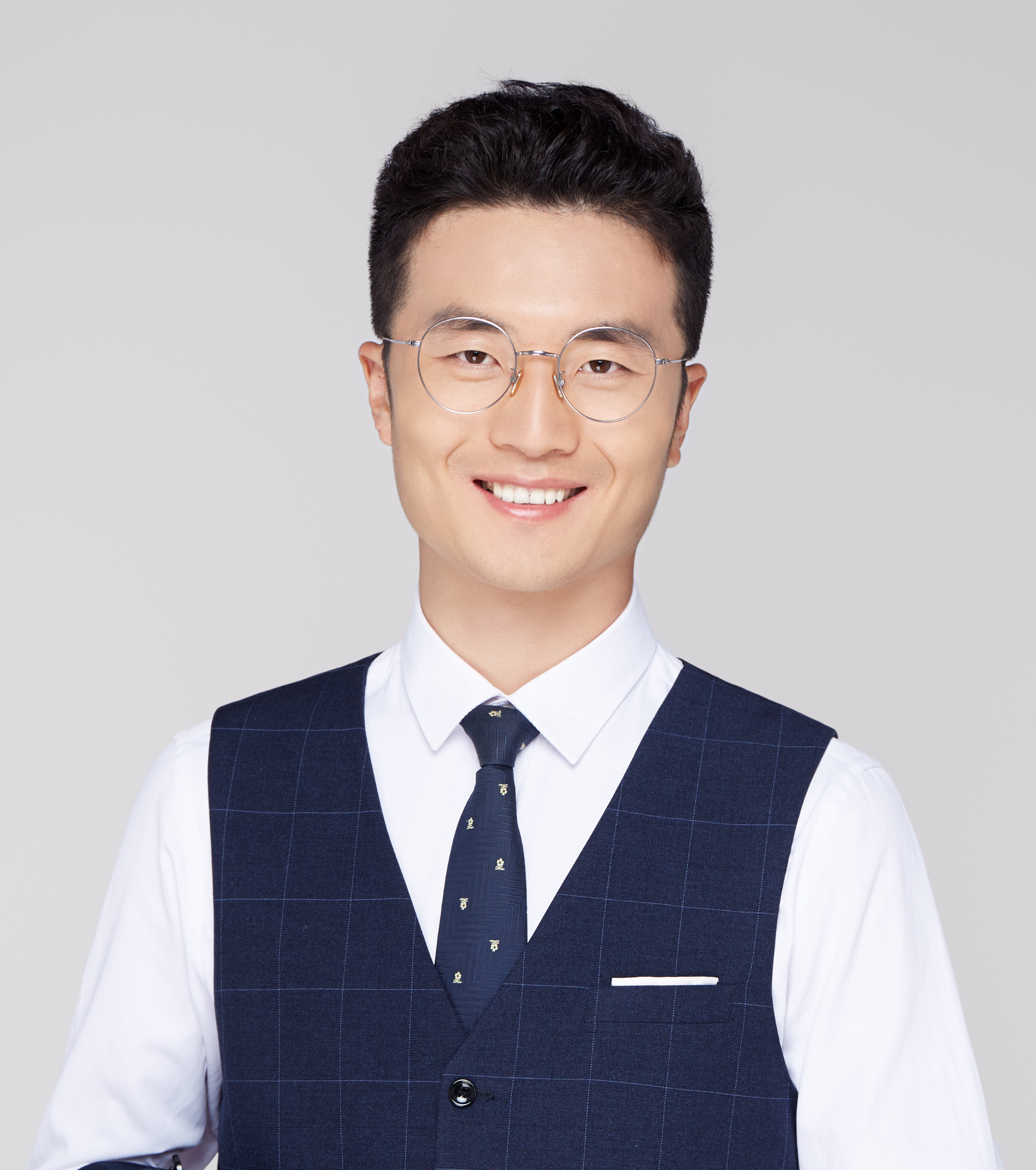}}]{Zhanxing Zhu} is Associate Professor at University of Southampton, UK, leading a research group on machine learning. He previously was senior research Professor at Changping National Lab and an assistant professor at Peking University. He obtained Ph.D degree in machine learning from University of Edinburgh. His research interests cover machine learning and its applications in various domains. Currently he mainly focuses on theoretical and methodological foundation of deep learning, and AI for Science. He has been recognized as the 2023 and 2024 AI 2000 Most Influential Scholar. 
\end{IEEEbiography}
\vspace{-11mm}
\begin{IEEEbiography}
[{\includegraphics[width=1in,height=1.3in,clip,keepaspectratio]{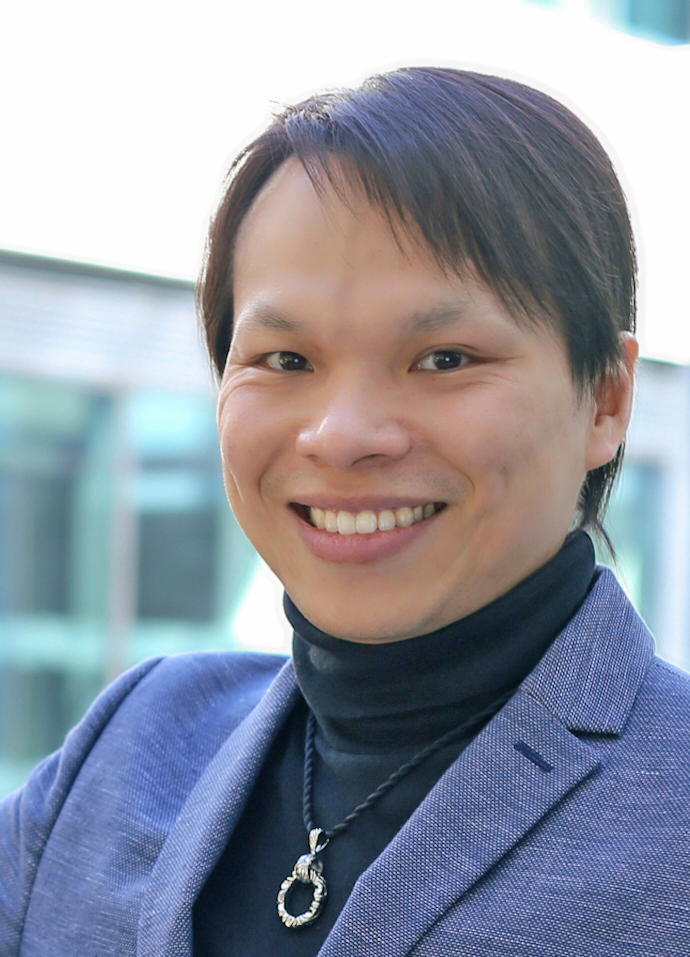}}]{Hubert P. H. Shum} (Senior Member, IEEE) is a Professor of Visual Computing and the Director of Research for the Department of Computer Science at Durham University, specialising in modelling spatio-temporal information with responsible AI. He is also a Co-Founder and the Co-Director of Durham University Space Research Centre. Before this, he was an Associate Professor at Northumbria University and a Postdoctoral Researcher at RIKEN Japan. He received his PhD degree from the University of Edinburgh. He chaired conferences such as Pacific Graphics, BMVC and SCA. He has authored over 200 research publications in the fields of Computer Vision, Computer Graphics and AI in Healthcare, underpinned by Responsible AI designs and algorithms.
\end{IEEEbiography}

\end{document}